\documentclass[letterpaper, 10pt, conference]{ieeeconf}
\usepackage{graphicx}
\usepackage{comment}
\usepackage{amsmath}
\usepackage{amssymb}
\usepackage{comment}
\usepackage{bm}
\usepackage{multirow}

\usepackage{color}
\usepackage{hyperref}
\usepackage[capitalise]{cleveref}
\usepackage{cite}

\DeclareMathOperator*{\argmax}{argmax}

\newif\ifproofread

\begin{document}
\proofreadtrue

\title{Epipolar-Guided Deep Object Matching for Scene Change Detection}

\author{
    \authorblockN{
        Kento Doi\authorrefmark{1}\authorrefmark{2},
        Ryuhei Hamaguchi\authorrefmark{2},
        Shun Iwase\authorrefmark{2},
        Rio Yokota\authorrefmark{2},
        Yutaka Matsuo\authorrefmark{1},
        Ken Sakurada\authorrefmark{2}
    }
    \authorblockA{\authorrefmark{1}
        The University of Tokyo, Tokyo, Japan}
    \authorblockA{\authorrefmark{2}
        National Institute of Advanced Industrial Science and Technology (AIST), Tokyo, Japan}
%    \authorblockA{\authorrefmark{3}Tokyo Insutitute of Technology, Tokyo, Japan}
}
% \author{
%     \authorblockN{Kento Doi}
%     \authorblockA{
%         The University of Tokyo,\\AIST
%         }
%     \and
%     \authorblockN{Ryuhei Hamaguchi}
%     \authorblockA{AIST}
%     \and
%     \authorblockN{Shun Iwase, Rio Yokota}
%     \authorblockA{Tokyo Institute of Technology}
%     \and
%     \authorblockN{Yutaka Matsuo}
%     \authorblockA{The University of Tokyo}
%     \and
%     \authorblockN{Ken Sakurada}
%     \authorblockA{AIST}
% }
% \author{Kento Doi, Ryuhei Hamaguchi, Shun Iwase, Rio Yokota, Yutaka Matsuo, and Ken Sakurada}

\maketitle

\begin{abstract}

This paper describes a viewpoint-robust object-based change detection network (OBJ-CDNet).
Mobile cameras such as drive recorders capture images from different viewpoints each time due to differences in camera trajectory and shutter timing.
However, previous methods for pixel-wise change detection are vulnerable to the viewpoint differences because they assume aligned image pairs as inputs.
To cope with the difficulty, we introduce a deep graph matching network that establishes object correspondence between an image pair.
The introduction enables us to detect object-wise scene changes without precise image alignment.
For more accurate object matching, we propose an epipolar-guided deep graph matching network (EGMNet), which incorporates the epipolar constraint into the deep graph matching layer used in OBJ-CDNet.
To evaluate our network's robustness against viewpoint differences, we created synthetic and real datasets for scene change detection from an image pair.
The experimental results verified the effectiveness of our network.

% \keywords{Change Detection, Object Detection, Graph Matching, Epipolar Geometry}

\end{abstract}

\section{Introduction\label{sec:introduction}}

Scene change detection has been exhaustively studied in the fields of computer vision and remote sensing for practical applications, such as anomaly detection, infrastructure inspection, and disaster prevention using images from satellites or surveillance cameras.
In recent years, the need for such research is increasing rapidly because maps must be kept up to date in applications of self-driving cars, augmented reality, and service robots. 
Specifically, for the navigation of autonomous vehicles, precise assessment of the latest landmarks is crucial, however, it is infeasible to manually update the maps for large city areas.
As an alternative, the previous studies \cite{wang2014cdnet,jhamtani2018learning,Huertas1998,bourdis2011constrained,crispell2012variable,Pollard2007,alcantarilla2018street,sakurada2015change} have presented methods using images taken by vehicle-mounted cameras to automatically detect the change regions.

Most of the methods require precise alignment to detect pixel-wise changes between input images taken at different times and cameras. 
In cases of satellite and surveillance cameras, viewpoint differences among images are generally small and easy to calibrate.
On the other hand, in cases of images captured by vehicle-mounted cameras, precise alignment is difficult since camera trajectory and shutter timing cannot be replicated across time.
Moreover, pixel-wise methods are vulnerable to illumination changes.

\begin{figure}
    \centering
    \includegraphics[width=0.95\hsize]{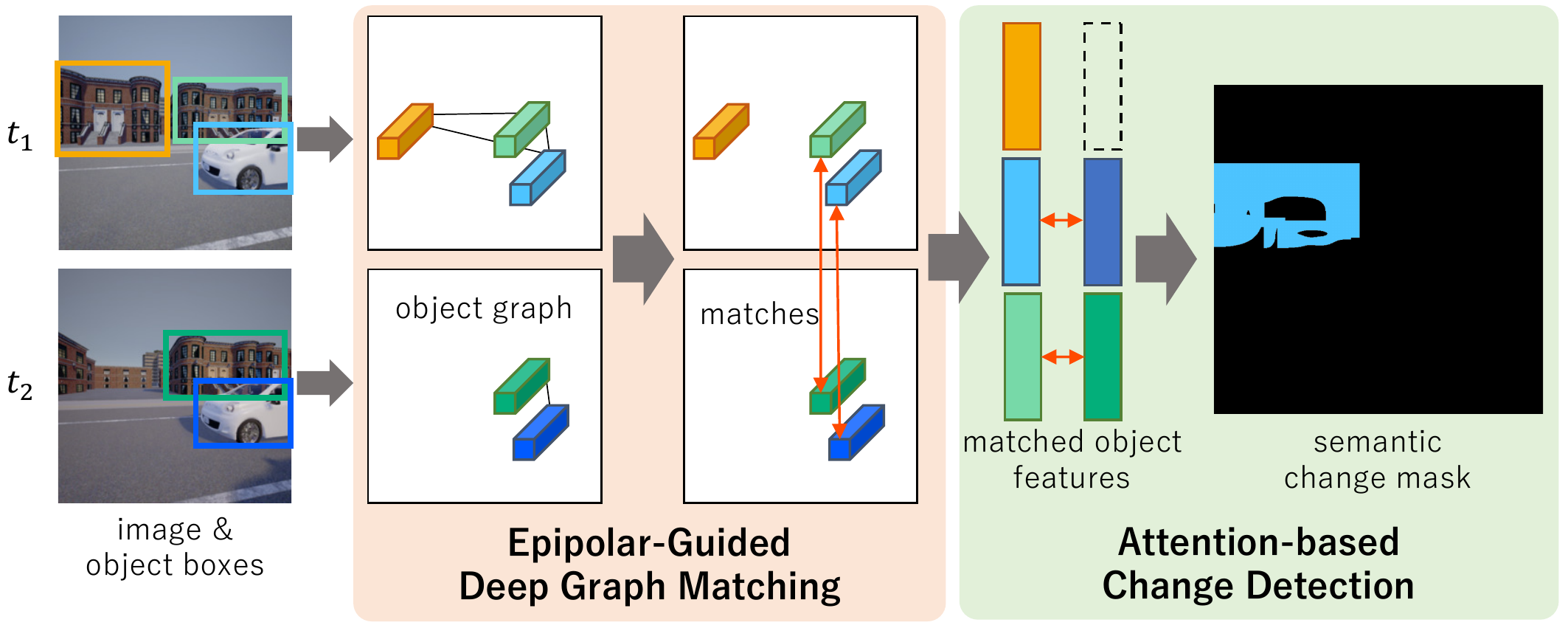}
    \caption{
        The overview of our object-based change detection network (OBJ-CDNet).
        Given an image pair and object detection boxes,
        we first establish object correspondences between images
        with deep graph matching which incorporates the epipolar constraint.
        Then, the attention-based network estimates the semantic change mask from matched object features.
        The OBJ-CDNet can learn the pixel-wise semantic change detection
        but requires only bounding boxes and correspondences of them as supervision.
    }
    \label{fig:overview}
\end{figure}

\begin{figure*}[!t]
    \centering
    \includegraphics[width=\hsize]{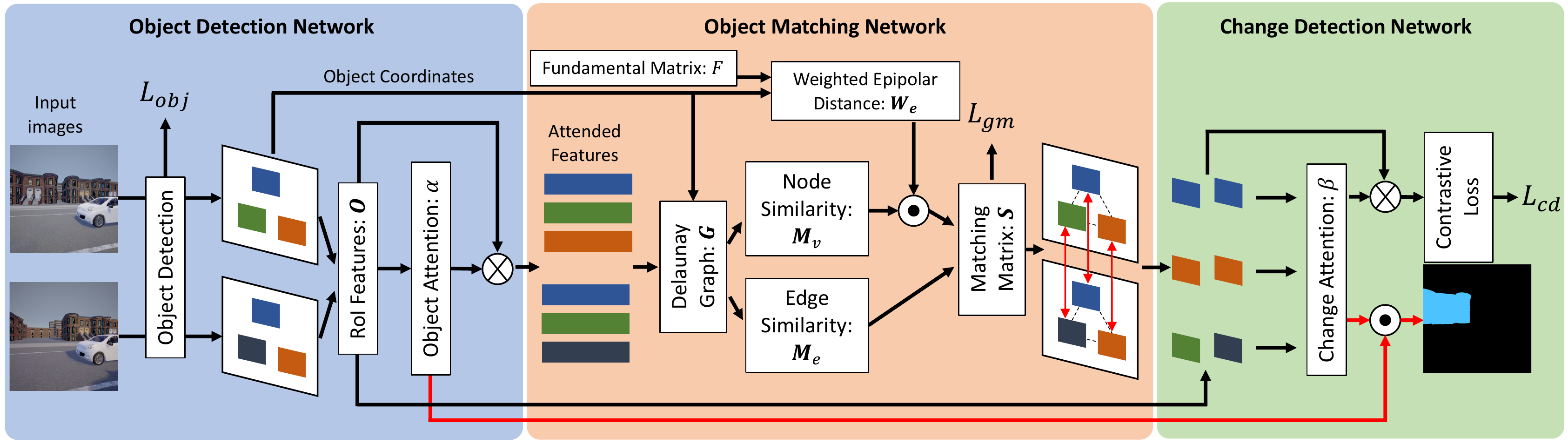}
    \caption{
        The details of our object-based change detection network (OBJ-CDNet).
        OBJ-CDNet is made up of three major components:
        the {\it object detection network} (\cref{subsec:objdet}),
        the {\it object matching network} (\cref{subsec:objmatch,subsec:epipolar}),
        and the {\it change detection network} (\cref{subsec:chgdet}).
        The first component takes an image pair as input and 
        predicts bounding boxes of objects in the scene.
        Then, object graphs are created from detected bounding boxes and
        image feature maps.
        The object matching layer creates an affinity matrix from
        the object graphs and a fundamental matrix and
        finds the optimal correspondences between objects.
        The final component uses an attention-based network
        to estimate a semantic change mask from the matched object features.
        }
    \label{fig:approach_details}
\end{figure*}

One possible solution to the difficulty of aligning images captured by vehicle-mounted cameras
is to perform depth estimation.
However, it is theoretically impossible to reconstruct the depth of the scene by stereo matching
from an image pair with scene changes.
Another possible strategy is the monocular depth estimation,
but it is difficult to accurately estimate the depth of thin objects such as streetlights and advertisement.
%To alleviate the difficulty, we propose an object-based change detection method with the deep graph matching and the epipolar-constraint to estimate pixel-wise scene changes.

To alleviate the difficulty, we propose an object-based change detection network (OBJ-CDNet) with deep graph matching (\cref{fig:approach_details}). Instead of precisely aligning input images, the OBJ-CDNet first extract object graphs from each image and applies deep graph matching to the graphs to detect object-wise scene changes. The method effectively incorporate epipolar constraint into the deep graph matching, which further improves robustness to viewpoint differences. By introducing an attention mechanism into the change detection network, the OBJ-CDNet can estimate pixel-wise changes without need for pixel-wise change annotations.
Furthermore, the OBJ-CDNet can be extended to estimate finer pixel-wise changes by replacing the attention module with a semantic segmentation network \cite{R-cnn2017MaskR-CNN} pre-trained with a semantic segmentation dataset, such as Cityscapes dataset \cite{Cordts2016Cityscapes}.
This extension can significantly reduce the amount of work required to create the dataset because labeling object-wise changes is much easier than pixel-wise ones while it achieves accuracy similar to fully-supervised methods that require pixel-wise change annotations.

Additionally, to evaluate our method, we created both synthetic and real datasets for object change detection, called CARLA- and GSV-OBJCD {\interfootnotelinepenalty=10000 \footnote{CARLA-OBJCD dataset and the annotaions of GSV-OBJCD dataset will be publicly available.}.
They consist of 15,000 and 500 scene perspective image pairs of CARLA \cite{Dosovitskiy17} and Google Street View images, respectively, with object bounding boxes and matching, the presence of change, pixel-level change map and object category annotation.}
To the best of our knowledge, CARLA-OBJCD dataset is the first publicly available large scale synthetic dataset for street-level scene change detection.

Our main contributions are as follows:
\begin{itemize}
    \item
        We propose an object-based change detection network utilizing deep graph matching
        that can also estimate pixel-wise changes with only object-wise change annotation.
    \item
        We propose the epipolar-guided deep graph matching network (EGMNet), in which the epipolar constraint is incorporated into a deep graph matching network.
    \item
        We built new synthetic and real datasets to benchmark scene change detection,
        which also contain object correspondences and pixel-level change masks between images.
\end{itemize}

This paper is organized as follows.
In \cref{sec:related_work}, we summarize the related work.
\Cref{sec:proposed_method} and \cref{sec:dataset} explain the details of
the proposed method and dataset respectively.
\Cref{sec:experiments} shows the experimental results
and finally, we conclude in \cref{sec:conclusion}.

\subsubsection*{Limitations}
Our approach can handle change detection where the objects can be clearly detected, but for ``stuff" \cite{panoptic_segmentation} where distinct objects cannot be clearly defined {\it e.g. pavement, walls, mountains}, object-based detection may not have the advantage we desire.
% However, since our current work focuses on scene change detection for vehicle-mounted cameras, the capability to handle ``stuff" that are typically not crucial for navigation is not a primary concern.
For tasks such as map update for autonomous driving, detecting changes of ``semantic objects" such as traffic signals and traffic signs, which indicate traffic rules, is more crucial than detecting changes of "stuff".
Therefore, this study focuses on the change of ``things".

\section{Related work\label{sec:related_work}}

\subsection{Change Detection}
Previous studies have proposed numerous scene change detection methods \cite{Radke2005,IbrahimEden2008,daudt2018fully} for images taken with surveillance cameras \cite{wang2014cdnet,jhamtani2018learning}, satellite and aerial imagery \cite{Huertas1998,bourdis2011constrained,crispell2012variable,Pollard2007}, and vehicular imagery \cite{BMVC2015_127,dong20174d}. 
Previous methods can be divided into two categories, detecting scene change of 2D (image space) or 3D space.
In the former category, methods in \cite{rosin1998thresholding,lopez2015features,wang2018m4cd} treat a 2D image of a surveillance camera as a background subtraction problem, and model the background image from multiple time periods.

Several methods that detect changes of 3D structures and their texture have also been proposed \cite{alcantarilla2018street,Schindler2010,Taneja2011,Taneja2013,Sakurada2013,MatzenECCV14}.
Most of these methods handle the scene depth as unique or probabilistic.
%by making use of Bayes' theorem or deep learning. 
They create the scene change model based on the idea, ``If there is a change in depth or texture, inconsistencies occur when projecting brightness values or features to images taken at different times".
When an input image pair has a viewpoint difference, pixel-level alignment and high-precision depth maps are necessary.

The recent success of convolutional neural networks (CNNs) enables more accurate and robust change detection \cite{sakurada2015change,daudt2018fully,zagoruyko2015learning,khan2017forest,guo2018learning}, semantic change detection \cite{daudt2018high,sakurada_icra2020} and change captioning \cite{jhamtani2018learning,park2019robust} in supervised, weakly-supervised \cite{khan2017learning,sakurada_icra2020}, semi-supervised \cite{Hamaguchi_2019_CVPR} and self-supervised \cite{furukawa2020iros} manners.
The CNNs drastically improve the robustness against changes of camera viewpoint and illumination condition between input images.
However, their estimation accuracies are greatly reduced if there is a viewpoint change beyond the range that can be considered with receptive fields of CNNs, because they estimate scene changes by comparing pixels or patches without the concept of objects.

As described in \cref{sec:introduction}, 
it is difficult to acquire a precisely aligned image pair
that is captured from a vehicle-mounted camera at different time points.
And it can reduce the accuracy of change detection methods.
The typical approach to solve this problem is to perform registration before change detection.
For example, Alcantarilla \textit{et al}. \cite{alcantarilla2018street}
proposed the change detection pipeline
that utilizes Simultaneous Localization and Mapping (SLAM) for the image registration.
After the alignment, they predict pixel-wise change mask
with convolutional neural network (CNN) using deconvolutional layers.
However, 3D reconstruction is computationally inefficient and
often inaccurate due to changes in the scene or the lack of images.
To address this issue, viewpoint-robust change detection methods
based on dense optical flow \cite{sakurada2017dense}, metric learning \cite{guo2018learning},
and correlation layer \cite{sakurada_icra2020} have been proposed ever.
The motivation of our work is the same as them,
but more robust to the large viewpoint difference and
requires lower annotation cost.

\subsection{Graph Matching}
Graph matching is a problem of finding a one-to-one node mapping between two graphs.
It is a fundamental task in computer vision which has many applications, such as key-point matching, template matching, structure from motion, multi-object tracking, and face recognition.
The original problem is classified as a quadratic assignment problem (QAP).
Since the QAP is known to be NP-hard, we often need to relax the problem in order to solve it directly.
Although there are some relaxation and optimization techniques \cite{Zhou2012,Zanfir_2018_CVPR} to achieve accurate matching, in this paper we incorporate the deep learning based graph matching method proposed by Zanfir and Sminchisescu \cite{Zanfir_2018_CVPR} to train object-level graph matching efficiently.
Different from the above methods, we leverage the epipolar constraint to perform more accurate deep graph matching.

\subsection{Attention-based Segmentation}
Zhou \textit{et al.} \cite{Zhou2016} proposed an attention-based object localization, which requires only image-level annotations for training.
They calculate a class activation map from weights and features in the last convolution layer with global average pooling,
and it localizes the salient region of an input image.
Seo \textit{et al.} propose progressive attention networks (PAN) \cite{Seo2016ProgressivePrediction} for attribute prediction, which successively applies attentive processes at multiple layers.
Jetly \textit{et al.} \cite{Jetley2018LEARNATTENTION} extends PAN to general classification networks,
and makes it possible to extract the different levels of attended semantic features. 
Their attention module can readily be incorporated into the object detection network by applying the same attention module not to a whole image but to region of interests (RoIs).
In our proposed network, we use attention maps to predict both pixel-wise object segmentation and change maps.

% ==========================================
% Proposed Method
% ==========================================
\section{Method\label{sec:proposed_method}}
%OBJ-CDNetは物体検出，物体グラフマッチング，変化検出の３つのネットワークで構成される(Figure??).
%OBJ-CDNet can detect object-level changes from an input image pair with different viewpoint. Furthermore, the method can estimate change mask by incorporating attention mechanisms into the framework, and can be trained using only object-level supervision.
As shown in \cref{fig:approach_details}, OBJ-CDNet consists of three parts: 1) object detection, 2) object graph matching, and 3) change detection network.
From objects detected by the object detection network, the object graph matching network achieves object-level matching through deep graph matching, and the change detection network then estimates changes using features from matched object pairs.
In the following, we first briefly introduce graph matching algorithm \cite{Zanfir_2018_CVPR} (\cref{subsec:preliminary}), then explain the object detection network (\cref{subsec:objdet}), the object graph matching network (\cref{subsec:objmatch}), and the change detection network (\cref{subsec:chgdet}).
In \cref{subsec:epipolar}, we explain how we can effectively incorporate epipolar constraint into graph matching algorithm.
%We first apply Faster R-CNN \cite{ren2015faster} with ResNet-101 \cite{he2016deep} to each image for object detection and the outputs (bounding boxes and their features) are fed into the object graph matching network directly.
%If a new object appears or disappears in the object graph matching network, it is immediately evaluated as a `Change' of type `No match'.
%If a different object is present in the same place, the object correspondence is first judged as a spatial `Match' from the positional relationship, then as a `Change' in the following change detection network.

% ==========================================
% Preliminary
% ==========================================
\subsection{Preliminary}
\label{subsec:preliminary}
Our object matching network is based on the deep graph matching algorithm proposed in \cite{Zanfir_2018_CVPR}. Below, we briefly introduce the algorithm.

Given two graphs $G_1 = (V_1, E_1)$, $G_2 = (V_2, E_2)$ with nodes $V_1$ and $V_2$ ($|V_1| = n$, $|V_2| = m$) and edges $E_1$ and $E_2$ ($|E_1| = p$, $|E_2| = q$), the objective of the graph matching algorithm is to find one-to-one mappings of nodes between two graphs based on affinity of node and edge features.
Let ${\bf x} \in \{0, 1\}^{nm}$ represent the mapping of nodes such that $x_{ij} = 1$ if $V_{1}^{i}$ is matched to $V_{2}^j$ and $x_{ij} = 0$ otherwise. Here, we denote $x_{ij}$ as $(i \times n + j)^{\text th}$ element of ${\bf x}$.
The symmetric affinity matrix, which holds the similarity between nodes and edges, is defined as ${\bf M} \in \mathbb{R}^{nm \times nm}$. ${\bf M}$ is originally a 4D tensor, which is contracted into a 2D matrix.
The diagonal components ${\bf M}_{(i_{1}i_{2},i_{1}i_{2})}$ depict the correlation of the $i_1^{\text th}$ node of $G_1$ and the $i_2^{\text th}$ node of $G_2$.
Furthermore, the non-diagonal components ${\bf M}_{(i_{1}i_{2},j_{1}j_{2})}$ depict the correlation of the edge $(i_{1}, j_{1}) \in E_1$  of $G_1$ and the edge $(i_{2}, j_{2}) \in E_2$ of $G_2$.
The graph matching problem can be formulated as follows:
\begin{equation}
    \begin{split}
        {\bf x}^* &= \underset{{\bf x}}{\argmax}
        \,\,\, {\bf x}^\mathrm{T} {\bf M} {\bf x} \\
        & s.t. \,\,\, {\bf C} {\bf x} \le {\bf 1}_{n+m},
        \quad \text{where} \quad {\bf C} =
        \begin{bmatrix}
            {\bf 1}_m^\mathrm{T} \otimes {\bf I}_n \\
            \,\,\, {\bf 1}_n \otimes {\bf I}_m^\mathrm{T} \\
        \end{bmatrix}
        .
    \end{split}
\end{equation}
Matrix ${\bf C}$ imposes one-to-one constraints.
This problem is a quadratic assignment problem and is NP-hard, which means we cannot solve it directly.
However, by introducing a unit constraint instead of a one-to-one constraint, we can solve it analytically.
The definition is as follows:
\begin{equation}
    {\bf x}^* = \underset{{\bf x}}{\argmax}  \,\,\,
    {\bf x}^\mathrm{T}{\bf M}{\bf x}  \quad s.t. \,\, {\bf x}^\mathrm{T} {\bf x} = 1.
    \label{eq:qap_relax}
\end{equation}
From this, we can achieve the solution ${\bf x}^*$ of \cref{eq:qap_relax} as the eigenvector corresponding to the maximum eigen value of ${\bf M}$.
By interpreting elements of ${\bf x}^{*}_{ij}$ as the reliability of $i \in V_1$ and $j \in V_2$, we can predict matchings.
Since correspondences are determined only by matrix $M$, the elements of matrix ${\bf M}$ can have a big effect on the accuracy of graph matching. Let the similarity of nodes be ${\bf M}_v \in \mathbb{R}^{n \times m}$, and the similarity of edges be ${\bf M}_e \in \mathbb{R}^{p \times q}$.
With the factorization method proposed in~\cite{Zhou2012}, we can obtain ${\bf M}$ from ${\bf M}_e$ and ${\bf M}_e$.

% ==========================================
% Object detection network
% ==========================================
\subsection{Object Detection Network}
\label{subsec:objdet}
In the object detection network, object bounding boxes are detected from each of the input images using Faster R-CNN \cite{ren2015faster} with ResNet-101 \cite{he2016deep}. From each detected bounding box, a feature volume ${\bf O}\in\mathbb{R}^{7 \times 7\times d}$ is obtained through RoI align. The RoI align is applied to the feature map before the last layer of the network. In order to aggregate features inside object regions, object attention maps are further estimated as in \cite{Jetley2018LEARNATTENTION}:
\begin{equation}
  {\bf o}_g = \text{GMP}\left(\text{conv} \left({\bf O} \right)\right) \in \mathbb{R}^d
\end{equation}

\begin{equation}
   c_{ij} = {\bf u}^T \left( {\bf o}_{ij} + {\bf o}_g  \right), \quad 
   \alpha_{ij} = \frac{\exp{(c_{ij})}}{\sum_{kl}\exp{(c_{kl})}}
\end{equation}
where GMP$(\cdot)$ is Global Max Pooling, $\text{conv}(\cdot)$ represents a convolutional layer, ${\bf u}$ are the weight parameters, and ${\bf o}_{ij}$ is the feature vector at $ij$-th spatial location in ${\bf O}$. The attention map ${\bf \alpha}$ is then used for estimating the change mask in the downstream change detection network.
In addition, ${\bf \alpha}$ is used for calculating the attended feature vector for the object as follows:
\begin{equation}
   {\bf \hat{o}}_{g} = \sum_{ij}\alpha_{ij}{\bf o}_{ij}
   \label{eq:obj_attention}
\end{equation}

As an alternative to the above attention based method, we also build the detection network based on pre-trained Mask R-CNN \cite{R-cnn2017MaskR-CNN},
and use the segmentation mask from its mask branch as a replacement of the attention map (OBJ-CDSegNet).
With the finer pixel-wise segmentation mask from Mask R-CNN, the downstream change detection network can precisely estimate change mask without using pixel-wise change annotations.
As we show in \cref{sec:experiments}, the OBJ-CDSegNet achieves competitive performance to the fully-supervised methods that require pixel-wise change annotations.
%As we show in \cref{sec:experiments}, the finer pixel-wise mask from Mask R-CNN brings large performance gain, making the proposed method competitive to the fully-supervised methods that require pixel-wise change annotations.

% ==========================================
% Object matching Network
% ==========================================
\subsection{Object Matching Network}
\label{subsec:objmatch}
\subsubsection{Approach}
The role of object matching network is to find one-to-one matching between detected objects from two input images.
In the network, object graphs $G_1(V_1, E_1)$, $G_2(V_2, E_2)$ are first constructed for each of the input images $I_1$ and $I_2$.
The detected objects are set as nodes $V_1$, $V_2$ ($|V_1| = n$, $|V_2| = m$), and the edges between nodes are constructed using the Delaunay triangulation ($|E_1| = p$, $|E_2| = q$).
The node features ${\bf X}_1\in\mathbb{R}^{n\times (d+2)}$ and ${\bf X}_2\in\mathbb{R}^{m\times (d+2)}$ are obtained for each node by concatenating the attended feature ${\bf \hat{o}}_g$ of the node and its coordinate.
As proposed in \cite{Yao2018ExploringVR}, the edge features ${\bf H_1}\in\mathbb{R}^{p\times d}$ and ${\bf H_2}\in\mathbb{R}^{q\times d}$ are taken from RoI pooling of the union bounding boxes of two joined nodes.
%instead of concatenating node features which belong to an edge as proposed in \cite{Zanfir_2018_CVPR}.

From the two graphs $G1$ and $G2$, the node affinity matrix ${\bf M}_p \in \mathbb{R}^{n \times m}$ is calculated as follows:
\begin{equation}
    {\bf M}_p = {\bf X}_1{\bf X}_2^T \odot {\bf W_{e}}
    \label{eq:node_affinity}
\end{equation}
where ${\bf W_{e}} \in \mathbb{R}^{n \times m}$ is a penalization term calculated by epipolar distance between objects. The term effectively incorporates epipolar constraint into the graph matching process; The matched objects should have small epipolar distance. The calculation of the term is explained in \cref{subsec:epipolar}.

Following \cite{Zanfir_2018_CVPR}, the edge affinity matrix ${\bf M}_e \in \mathbb{R}^{p \times q}$ is calculated as follows: 
\begin{equation}
    {\bf M}_e = {\bf H}_1{\bf \Lambda}{\bf H}_2^T
    \label{eq:edge_affinity}
\end{equation}
where $\Lambda\in\mathbb{R}^{d\times d}$ is a block symmetric parameter matrix.

From the affinity matrices, we can obtain matching matrix ${\bf S}\in\mathbb{R}^{n\times m}$ using deep graph matching algorithm \cite{Zanfir_2018_CVPR} as:
\begin{equation}
    {\bf S} = \text{GraphMatch}({\bf M}_p, {\bf M}_e)
\end{equation}
The $ij^{th}$ element of ${\bf S}$ represents matching confidence between $i^{th}$ node in $V_1$ and $j^{th}$ node in $V_2$.

It is common to utilize the bi-stochastic constraint -- a method to convert the output to a doubly stochastic matrix to improve the accuracy in complete matching problems.
However, in our proposed method we allow nodes with no match, thus the constraint is inappropriate, so we do not apply this method. 
% We compare the change in accuracy with and without enforcing the bi-stochastic constraint in the experimental section.

\subsubsection{Training}
During training, a graph matching loss is calculated as a cross entropy loss between the estimated matching matrix ${\bf S}$ and the ground truth matching matrix ${\bf S}^{gt}$. In the loss calculation, ${\bf S}$ is interpreted as a probability matrix in the following two ways:
\begin{equation}
  {\bf R}^{1\rightarrow 2} = \sigma_{row} \left( {\bf S} \right), \quad
  {\bf R}^{2\rightarrow 1} = \sigma_{col} \left( {\bf S} \right)
\end{equation}
where $\sigma_{row}$ and $\sigma_{col}$ represent row- and col-wise softmax operator respectively. ${\bf R}^{1\rightarrow 2}$ and ${\bf R}^{2\rightarrow 1}$ is the matching probability from nodes $V_1$ to nodes $V_2$ and from nodes $V_2$ to nodes $V_1$, respectively. The matching loss function can then be as follows:
\begin{equation}
  L_{gm}({\bf S}, {\bf S}^{gt}) =  \sum_{ij} \left( S_{ij}^{gt}\log {R}_{ij}^{1\rightarrow 2} + S_{ij}^{gt}\log {R}_{ij}^{2\rightarrow 1} \right)
\end{equation}

\begin{comment}
During training, we are given ground truth matching ${\bf S}^{gt}$. From ${\bf S}^{gt}$ and the estimated confidence matrix ${\bf S}$, we can respectively calculate the true coordinate of the the matched node ${\bf D}^{gt}\in\mathbb{R}^{n\times 2}$ and the estimated coordinate of the matched node ${\bf D}\in\mathbb{R}^{n\times 2}$:
\begin{equation}
  {\bf R} = \text{softmax} \left( {\bf S}, \tau \right), \quad
  {\bf D} = {\bf R} {\bf P}_2
\end{equation}
\begin{equation}
  {\bf D}^{gt} = {\bf S}^{gt} {\bf P}_2
\end{equation}
where $\text{softmax} \left( \cdot, \tau \right)$ represent row-wise softmax operation with temparature $\tau$,
and ${\bm P_{2}}$ is the spatial positions of nodes in ${\bm V_{2}}$.
From the coordinates, the following loss function can be defined as graph matching loss:
\begin{equation}
  L_{gm}({\bf S}, {\bf S}^{gt}) =  \sum_i^n \left\lVert {\bf D}_{i} - {\bf D}_{i}^{gt} \right\rVert_2 \\
\end{equation}
\end{comment}

\subsubsection{Inference}
We evaluate the existence of matching at the time of inference, based on the attained matching matrix ${\bf S}$.
The one-to-one matching can be established in the following function:
\begin{equation}
  {\bf S} \, _{ij} = \left\{ \begin{array}{ll}
      1 & (\text{if} \quad \underset{t}{\argmax} \, {{\bf S}_{it}} = j \,\, \text{and} \\
        & \underset{t}{\argmax} \, {{\bf S}_{tj}} = i\,\, \text{and} \,\, {\bf S}_{ij} > \gamma) \\
    0 & (\text{otherwise})
  \end{array} \right.
\label{eq:matching_inf}
\end{equation}
where $\gamma$ is a hyperparameter that determines the minimum confidence required for the matching.
A node $i \in G_1$ and a node $j \in G_2$ are considered as a match if ${\bf S} \, _{ij} = 1$.
If no node is matched, that node is labeled as `not matched'. 

For objects labeled as `not matched', they are immediately classified as `changed', and for objects with matched pair, they are further processed in the succeeding step using change detection network. 

% ==========================================
% Epipolar-guided deep graph matching network
% ==========================================
\subsection{Epipolar-guided Deep Graph Matching Network}
\label{subsec:epipolar}
The key idea behind the epipolar-guided deep graph matching network (EGMNet) is to improve the ability to distinguish matching and non-matching objects based on the epipolar constraint between two images in the presence of large viewpoint differences and object detection errors. The constraint can be effectively incorporated into the graph matching network by penalizing the node affinity using epipolar distance between objects (\cref{eq:node_affinity}). The details of EGMNet are described below.

Epipolar distances between the objects are calculated with a fundamental matrix ${\bf F}$.
% Epipolar distances between the objects are calculated with an fundamental matrix ${\bf F}$, which is estimated prior to this process from Nister's five-point algorithm \cite{nister2004efficient} with SIFT features \cite{lowe2004distinctive}.
At this point, as it is believed that the larger the rectangular area becomes, the greater the variance of the center coordinate is from the epipolar line, we introduce {\it normalized epipolar distance} which considers the rectangular area.
Here we consider the epipolar distance of $i \in V_1$ and $j \in V_2$.
Each coordinate of the node can be written as ${\bf p}_{i}$ and ${\bf p}_{j}$, hence the epipolar line as ${\bf l}_j = {\bf F} {\bf p}_j$.
Therefore, the perpendicular vector ${\bf v}_{ij}$ from ${\bf p}_{i}$ to the epipolar line ${\bf l}_j$ is described as follows.
\begin{equation}
  {\bf v}_{ij} = \frac{|{\bf p}_i {\bf F} {\bf p}_j|}{\left\lVert \, {\bf l}_j [\,:2\,] \, \right\rVert} \, {\bf t}
\end{equation}
where ${\bf x}[\, \text{:}2 \,]$ is a shorthand to represent the first two elements of vector ${\bf x}$ and ${\bf t}$ is the unit vector in direction ${\bf v}_{ij}$.
% ノード$i$に対する矩形領域の高さ, 幅をそれぞれ$h_i, w_i$と表すと, 重み付きエピポーラ距離${\bf D} \in \mathbb{R}^{n \times m}$を次のように定義する.
Let the height and width of the rectangular area of node $j$ be $h_j, w_j$. Then, the normalized epipolar distance matrix ${\bf D} \in \mathbb{R}^{n \times m}$ is calculated as follows.
\begin{equation}
  % D_{ij} = \sqrt{
  %   \left(\frac{x_{ij}}{w_i}\right)^2 +
  %   \left(\frac{y_{ij}}{h_i}\right)^2
  % }
  {\bf D}_{ij} = \sqrt{
    {\bf v}_{ij}^\mathrm{T}
    \begin{bmatrix}
      \frac{1}{{w_j}^2} & 0 \\
      0 & \frac{1}{{h_j}^2}
    \end{bmatrix}
    {\bf v}_{ij}
  }
  \label{eq:epi_dist}
\end{equation}
%\begin{equation}
%  \sigma_x = w_i, \quad \sigma_y = h_i
%  \label{eq:epi_dist_variances}
%\end{equation}
% 矩形領域が大きければ大きいほど, 矩形領域の相対的な位置ずれがエピポーラ距離に与える影響が大きくなる.
% そこで, その影響を緩和するために矩形領域の大きさを分散として用いる.
%The larger the bounding box, the greater the influence of the relative displacement on the epipolar distance.
%Therefore, in order to mitigate it, we use the width and height of the bounding box as variances.
% 上式で得た距離を重みとして用いることによって, 次のエピポーラ重み付きノード間類似度を得る.
We finally attain epipolar guided node similarities
\begin{equation}
  {\bf M}_p = {\bf X}_1 {\bf X}_2^\mathrm{T} \odot \text{exp}\left(- \frac{{\bf D} \, ^2}{2 \, \mu_{\bf D}^2}\right)
  \label{eq:epipolar_similarity}
\end{equation}
where $\mu_{{\bf D}}$ is the standard deviation of ${\bf D}$.
% この操作により, 重み付きエピポーラ距離が大きいペアの類似度を小さくしている.
With this operation, we minimize similarities of pairs which have large normalized epipolar distance.
% ${\bf M}_e$に関しては, []と同様の手法でパラメタ化を行う.

% =================================
% Change detection network
% =================================
\subsection{Change Detection Network}
\label{subsec:chgdet}
For each matched object pair obtained from object matching network,
the change detection network further classifies if there is a change between objects, and at the same time estimate the change mask.
In order to attend to changed region, the change attention maps ${\bf \beta}_1, {\bf \beta}_2\in\mathbb{R}^{7\times 7}$ are estimated using RoI feature volumes ${\bf O}_{1}$ and ${\bf O}_{2}$ of a matched object pair:
\begin{equation}
    {\bf O}_{\text{diff}} = {\bf O}_{1} - {\bf O}_{2}
\end{equation}
\begin{equation}
    {\bf \beta}_{1} = \sigma \left( \text{conv}_2 \left( \text{conv}_1 \left(
        [{\bf O}_{1}; \, {\bf O}_{\text{diff}}] \right) \right)  \right)
\end{equation}
\begin{equation}
    {\bf \beta}_{2} = \sigma \left( \text{conv}_2 \left( \text{conv}_1 \left(
        [{\bf O}_{2}; \, {\bf O}_{\text{diff}}] \right) \right)  \right)
\end{equation}
where $\sigma$ is a softmax operator along the spatial dimensions. From the change attention maps, aggregated feature vector for attended region is obtained:
\begin{equation}
  {\bf c}_{1} = \sum_{ij}\beta_{1,ij}{\bf O}_{1,ij}
\end{equation}
\begin{equation}
  {\bf c}_{2} = \sum_{ij}\beta_{2,ij}{\bf O}_{2,ij}
\end{equation}
During training, we train the feature space using contrastive loss \cite{contrastive_loss} between ${\bf c}_{1}$ and ${\bf c}_{2}$:
\begin{equation}
    d = \left\lVert {\bf c}_{1} - {\bf c}_{2} \right\rVert_2^2, \quad
    L_{cd} = td + (1-t)max(\tau_m - d, 0)
\label{eq:contrastive_loss}
\end{equation}
where $t$ is a ground truth label ($t=0$ represent `changed' and $t=1$ represent `not changed') and $\tau_m$ is a margin parameter for the contrastive loss.
Once we train the feature space, we can detect change based on the distance between attended features ${\bf c}_1$ and ${\bf c}_2$

Finally, we can also estimate the change masks from the change attention maps and the object attention maps from the object detection network:
\begin{equation}
    {\bf m_1} = {\bf \alpha}_1 \odot {\bf \beta}_1, \quad
    {\bf m_2} = {\bf \alpha}_2 \odot {\bf \beta}_2
\end{equation}

\subsection{Loss Function}
We train the three networks shown in \cref{fig:approach_details} at the same time.
We define the loss function as the summation of their loss functions, $L_{obj}$, $L_{gm}$, and $L_{cd}$ where $L_{obj}$ is the loss function used in Faster R-CNN \cite{ren2015faster}.
We do not use any weight parameters for three loss terms.

% ==========================
% Datasets
% ==========================
\section{Datasets\label{sec:dataset}}
% 変化の定義とかかく
Although there are several publicly available change detection datasets, such as CD2014 \cite{wang2014cdnet}, PCD2015 \cite{sakurada2015change} and VL-CMU-CD \cite{alcantarilla2018street} which contain image pairs and change maps, there is no dataset which contains paired object-level annotations, such as bounding boxes and categories.
Therefore, we built synthetic and real datasets for object change detection, called CARLA- and GSV-OBJCD, to evaluate our model and facilitate new researches on both object- and pixel-level scene change detection.

\begin{figure}[t]
  \centering
  \includegraphics[width=\hsize]{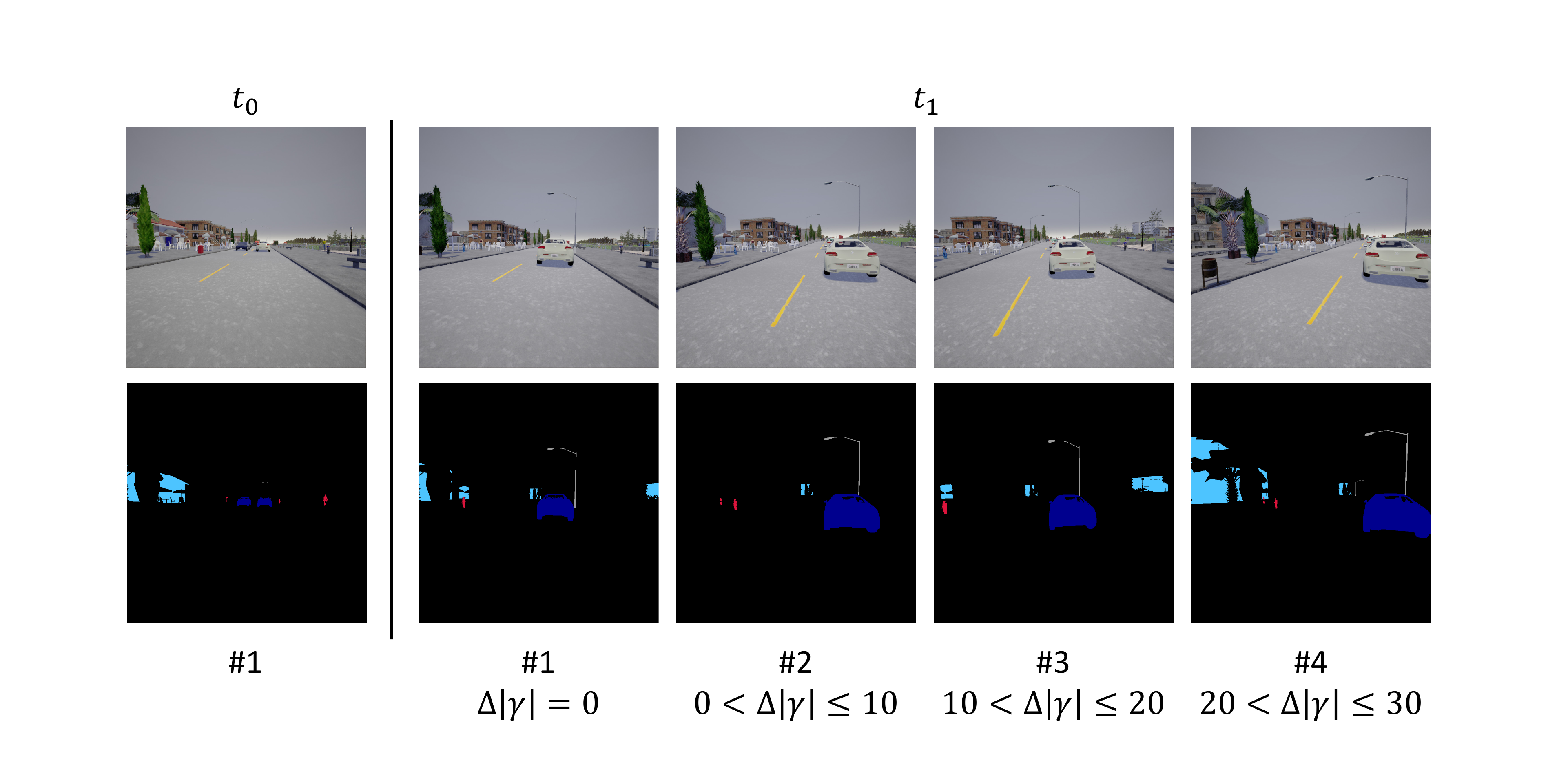}
  \caption{Examples from the CARLA-OBJCD dataset. There are four pair images per location which have different camera viewpoints.}
  \label{fig:example_carla}
\end{figure}

\subsection{CARLA-OBJCD Dataset}
\label{subsec:carla_dataset}

This dataset contains 15,000 scene perspective image pairs and their object- and pixel-wise semantic change annotations including object correspondences between images, which are automatically generated under different illumination settings with the CARLA simulator \cite{Dosovitskiy17}. For evaluation of robustness against camera viewpoint difference, we prepared four variant sets, each of which has 15,000 scenes with different ranges of camera viewpoint difference. 
The camera pose of the reference image is the same across the four sets, and that of the respective target image varies for the four sets. In set \#1, there is no viewpoint difference between reference and target. In sets \#2, \#3, \#4, relative poses of horizontal translation $\Delta x, \Delta y$ [m], roll $\Delta \alpha$ and pitch $\Delta \beta$ [deg] are sampled from the same range $-1 \leq \Delta x, \Delta y \leq 1,  -5 \leq \Delta \alpha, \Delta \beta \leq 5$. Only yaw $\gamma$ [deg] is sampled from different ranges, \#2: $0$ $\leq \Delta |\gamma| \leq$ $10$, \#3: $10$ $\leq \Delta |\gamma| \leq$ $20$, and \#4: $20$ $\leq \Delta |\gamma| \leq$ $30$.
We showed the examples of CARLA-OBJCD dataset in \cref{fig:example_carla}.

\subsection{GSV-OBJCD Dataset}
\label{subsec:gsv_dataset}

This dataset contains 500 pairs of perspective images cropped from Panoramic Semantic Change Detection (PSCD) dataset \cite{sakurada_icra2020} and manually added annotations of the same type as those in the CARLA-OBJCD dataset (i.e. pixel- and object-wise semantic change labels and matching correspondences of 10388 objects).
In contrast to other change detection datasets \cite{alcantarilla2018street,sakurada2015change}, it is relatively challenging because of the large viewpoint differences and multiple small changes.
We believe that the ability to deal with such difficulties is necessary for real applications. 

\begin{table}[t]
    \caption{
        Object matching accuracy for CARLA- and GSV-OBJCD datasets.
        Dataset \#1 has no viewpoint difference and \#4 has the largest ones.
        FC and DT represent fully connected and delaunay triangulation, respectively.
        }
    \begin{center}
        \begin{tabular}{c|cccc|c}
            \multirow{2}{*}{}& \multicolumn{4}{c|}{CARLA} & \multirow{2}{*}{GSV} \\ \cline{1-5}
            & \#1 & \#2 & \#3 & \#4 &   \\ \hline
            NN & {\bf 0.770} & 0.645 & 0.644 & 0.581 & 0.777  \\ \hline
            GMN \cite{Zanfir_2018_CVPR}  & 0.756 & 0.665 & 0.666 & 0.625 & 0.791  \\ \hline\hline
            ENN & 0.761 & 0.662 & 0.662 & 0.643 & 0.804  \\ \hline
            EGMNet (FC) & 0.751 & 0.738 & 0.736 & 0.753 & 0.809   \\ \hline
            EGMNet (DT) & 0.760 & {\bf 0.746} & {\bf 0.747} &  {\bf 0.760} & {\bf 0.822}   \\ \hline
            \hline
        \end{tabular}
    \end{center}
    \label{tab:graph_matching_result}
\end{table}

\begin{comment}
\begin{table}[t]
    \caption{
        Object matching accuracy for CARLA- and GSV-OBJCD datasets.
        Dataset \#1 has no viewpoint difference and \#4 has the largest ones.
        FC and DT represent fully connected and delaunay triangulation, respectively.
        }
    \begin{center}
    \scalebox{0.9}{
        \begin{tabular}{c|cc|cccc|c}
            & \multirow{2}{*}{\hspace{-1mm}Graph} & \multirow{2}{*}{\hspace{-3mm}Epipolar\hspace{-1mm}} & \multicolumn{4}{c|}{CARLA} & \multirow{2}{*}{GSV} \\ \cline{4-7}
            &&& \#1 & \#2 & \#3 & \#4 &   \\ \hline
            NN &&& {\bf 0.770} & 0.645 & 0.644 & 0.581 & 0.777  \\ \hline
            GMN \cite{Zanfir_2018_CVPR}  &\checkmark&& 0.756 & 0.665 & 0.666 & 0.625 & 0.791  \\ \hline
            ENN &&\checkmark& 0.761 & 0.662 & 0.662 & 0.643 & 0.804  \\ \hline
            EGMNet (FC) &\checkmark&\checkmark& 0.751 & 0.738 & 0.736 & 0.753 & 0.809   \\ \hline
            EGMNet (DT) &\checkmark&\checkmark& 0.760 & {\bf 0.746} & {\bf 0.747} &  {\bf 0.760} & {\bf 0.822}   \\ \hline
        \end{tabular}
        }
    \end{center}
    \label{tab:graph_matching_result}
\end{table}
\end{comment}

\begin{table*}[t]

    \begin{minipage}{0.65\hsize}
        \caption{
            mIoU of change detection for CARLA-OBJCD dataset.
            Dataset \#1 has no viewpoint difference and \#4 has the largest ones.
        }
            \begin{tabular}{c|cccc}
                -- Pixel-based method (FS) -- & \#1 & \#2 & \#3 & \#4 \\ \hline\hline
                CDNet \cite{sakurada2017dense} & {\bf 0.754} & 0.602 & 0.578 & 0.554 \\ \hline
                CosimNet-3layer-l2 \cite{guo2018learning} & 0.693 & {\bf 0.610} & {\bf 0.612} & {\bf 0.616} \\ \hline
                CSCDNet \cite{sakurada_icra2020} & {\bf 0.754} & 0.600 & 0.605 & 0.560 \\ \hline\hline
                -- OBJ-CDNet (Att) -- & \#1 & \#2 & \#3 & \#4 \\ \hline\hline
                NN & 0.552 (0.625) & 0.539 (0.608) & 0.539 (0.606) & 0.535 (0.599) \\ \hline
                ENN & 0.534 (0.581) & 0.518 (0.568) & 0.520 (0.569) & 0.518 (0.570) \\ \hline
                GMN \cite{Zanfir_2018_CVPR} & 0.566 (0.637) & 0.546 (0.619) & 0.554 (0.619) & {\bf 0.542} (0.607) \\ \hline
                %-- Ours -- &\rule[0mm]{0mm}{3mm} \\ \hline
                EGMNet & {\bf 0.577} (0.632) & {\bf 0.531} (0.630) & {\bf 0.584} (0.630) & 0.533 (0.619) \\ \hline \hline
                -- OBJ-CDSegNet (Seg) -- & \#1 & \#2 & \#3 & \#4 \\ \hline\hline
                NN & 0.594 (0.661) & 0.599 (0.604) & 0.597 (0.666) & 0.590 (0.656) \\ \hline
                ENN & 0.595 (0.663) & 0.597 (0.658) & 0.597 (0.661) & 0.598 (0.668) \\ \hline
                GMN & {\bf 0.599} (0.683) & 0.598 (0.680) & 0.596 (0.678) & 0.597 (0.672) \\ \hline
                %-- Ours -- &\rule[0mm]{0mm}{3mm} \\ \hline
                EGMNet & 0.596 (0.683) & {\bf 0.600} (0.684) & {\bf 0.600} (0.683) & {\bf 0.605} (0.683)  \\ \hline \hline
            \end{tabular}
        \label{tab:change_detection_pixel_carla_result}
    \end{minipage}
    \hfill
    \begin{minipage}{0.325\hsize}
        \centering
        \caption{
            mIoU of change detection for GSV-OBJCD dataset.
        }
        \begin{tabular}{c|c}
            -- Pixel-based method (FS) -- & \\ \hline\hline
            CDNet \cite{sakurada2017dense} & 0.556 \\ \hline
            CosimNet-3layer-l2 \cite{guo2018learning} & 0.547  \\ \hline
            CSCDNet \cite{sakurada_icra2020} & {\bf 0.583} \\ \hline\hline
            -- OBJ-CDNet (Att) -- & \\ \hline\hline
            NN & 0.471 (0.564)\\ \hline
            ENN & 0.482 (0.566) \\ \hline
            GMN & 0.483 (0.567) \\ \hline
            EGMNet & {\bf 0.492} (0.576) \\ \hline \hline
            -- OBJ-CDSegNet (Seg) -- & \\ \hline\hline
            NN & 0.556 (0.589) \\ \hline
            ENN & 0.559 (0.587) \\ \hline
            GMN & 0.558 (0.589) \\ \hline
            EGMNet & {\bf 0.563} (0.593) \\ \hline \hline
        \end{tabular}
        \label{tab:change_detection_pixel_gsv_result}
    \end{minipage}

\end{table*}

\begin{figure*}
    \begin{minipage}{0.02\hsize}
        \raggedright
        {\footnotesize (a)}
    \end{minipage}
    \begin{minipage}{0.1125\hsize}
        \centering
        \includegraphics[width=0.95\hsize, keepaspectratio]{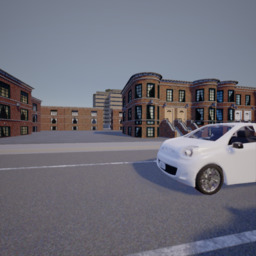}
    \end{minipage}
    \begin{minipage}{0.1125\hsize}
        \centering
        \includegraphics[width=0.95\hsize, keepaspectratio]{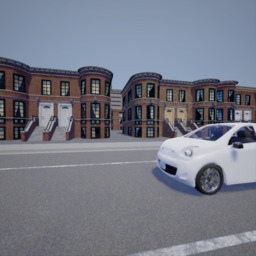}
    \end{minipage}
    \begin{minipage}{0.1125\hsize}
        \centering
        \includegraphics[width=0.95\hsize, keepaspectratio]{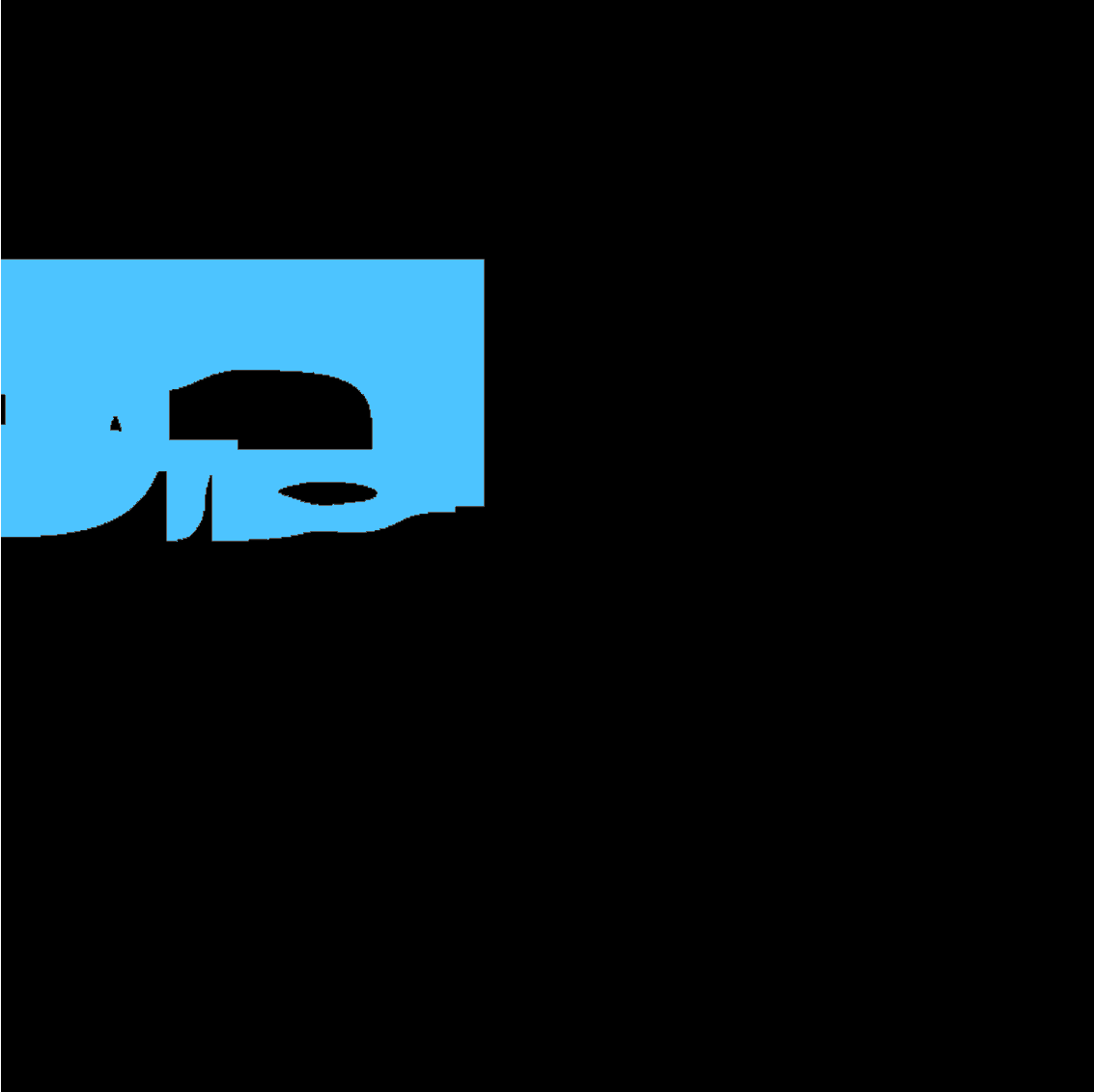}
    \end{minipage}
    \begin{minipage}{0.1125\hsize}
        \centering
        \includegraphics[width=0.95\hsize, keepaspectratio]{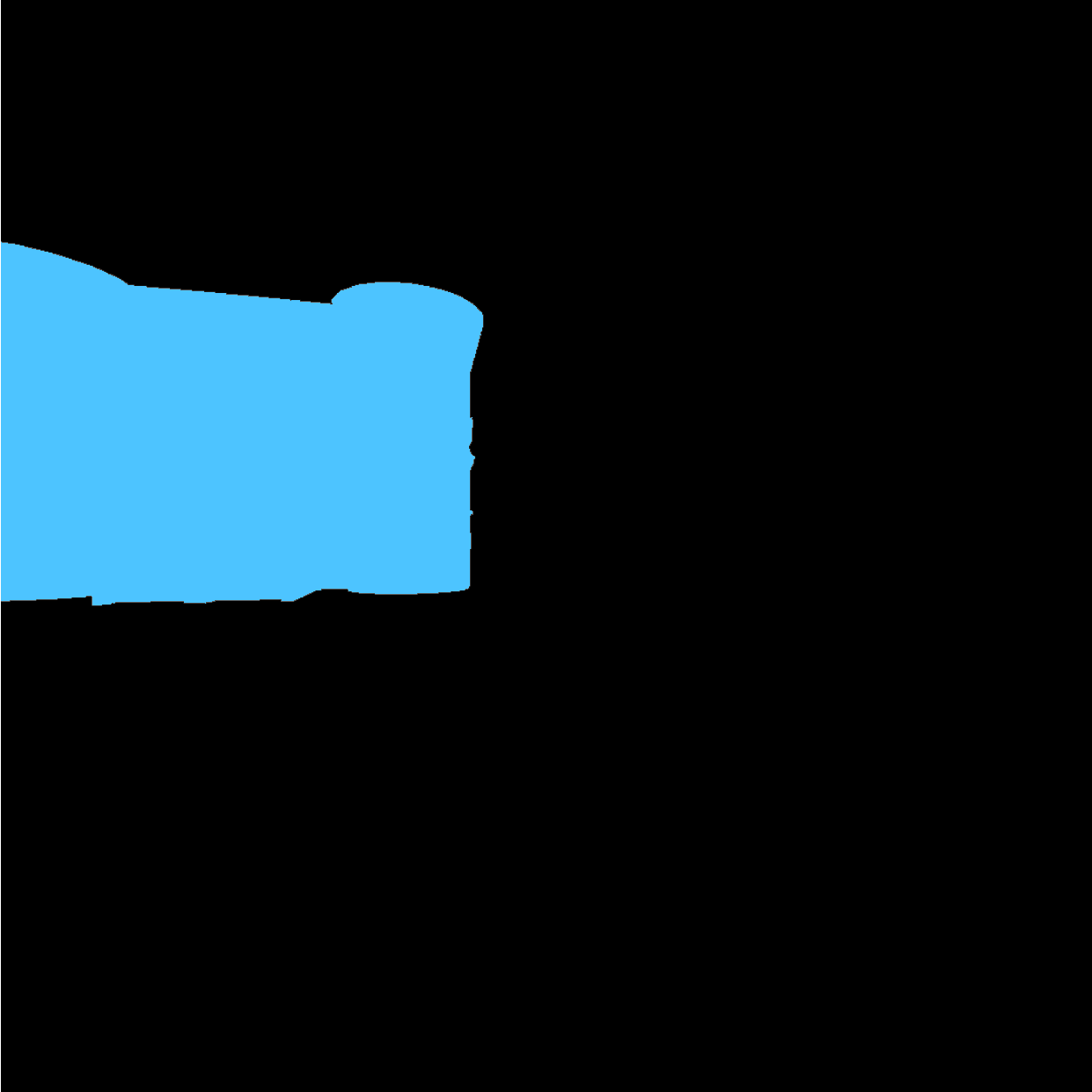}
    \end{minipage}%
    \hfill
    \begin{minipage}{0.02\hsize}
        \raggedright
        {\footnotesize (c)}
    \end{minipage}
    \begin{minipage}{0.1125\hsize}
        \centering
        \includegraphics[width=0.95\hsize, keepaspectratio]{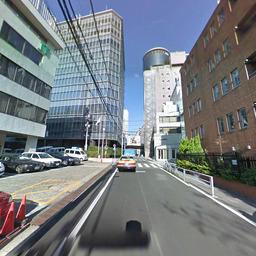}
    \end{minipage}
    \begin{minipage}{0.1125\hsize}
        \centering
        \includegraphics[width=0.95\hsize, keepaspectratio]{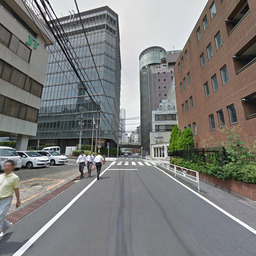}
    \end{minipage}
    \begin{minipage}{0.1125\hsize}
        \centering
        \includegraphics[width=0.95\hsize, keepaspectratio]{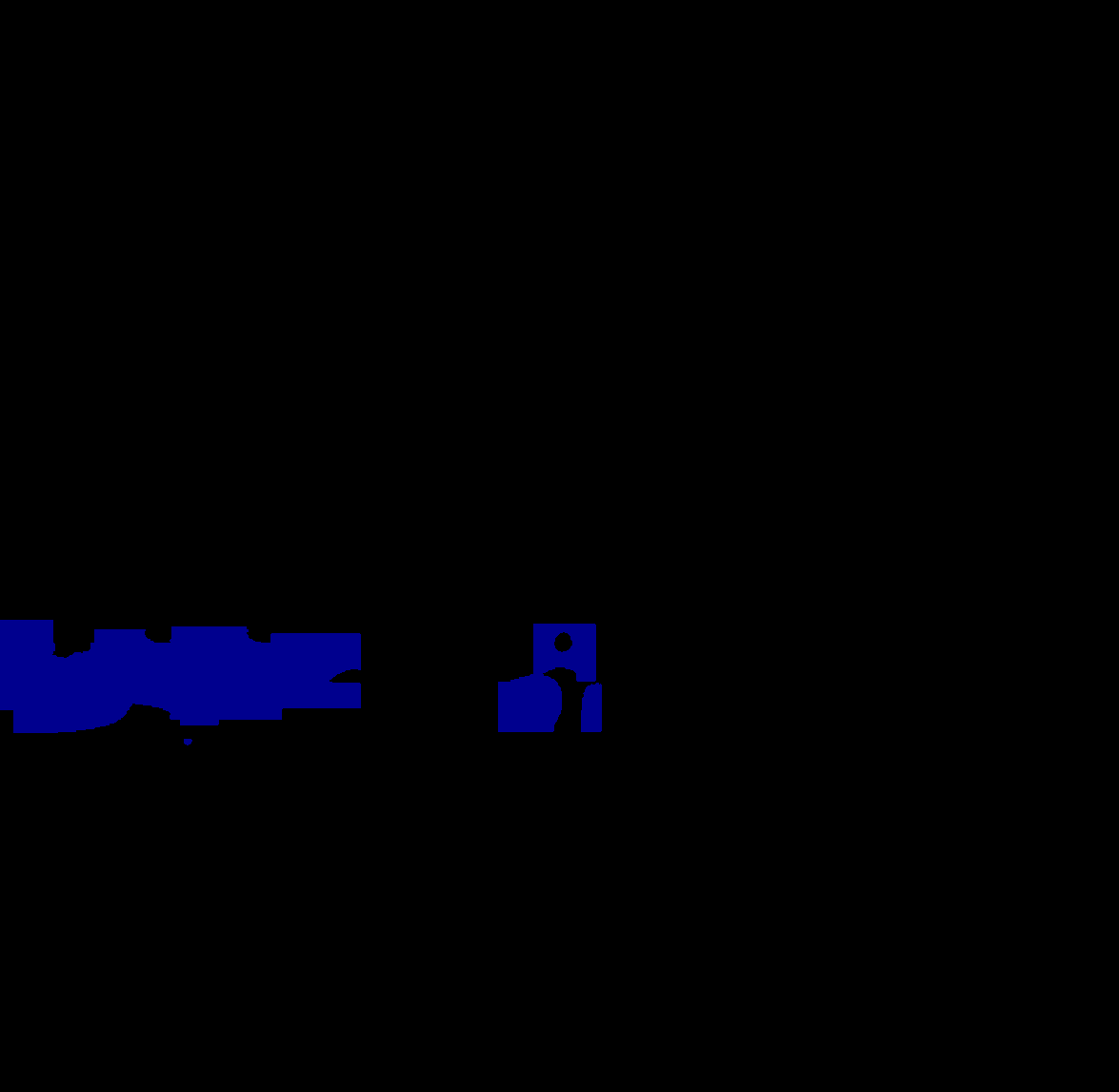}
    \end{minipage}
    \begin{minipage}{0.1125\hsize}
        \centering
        \includegraphics[width=0.95\hsize, keepaspectratio]{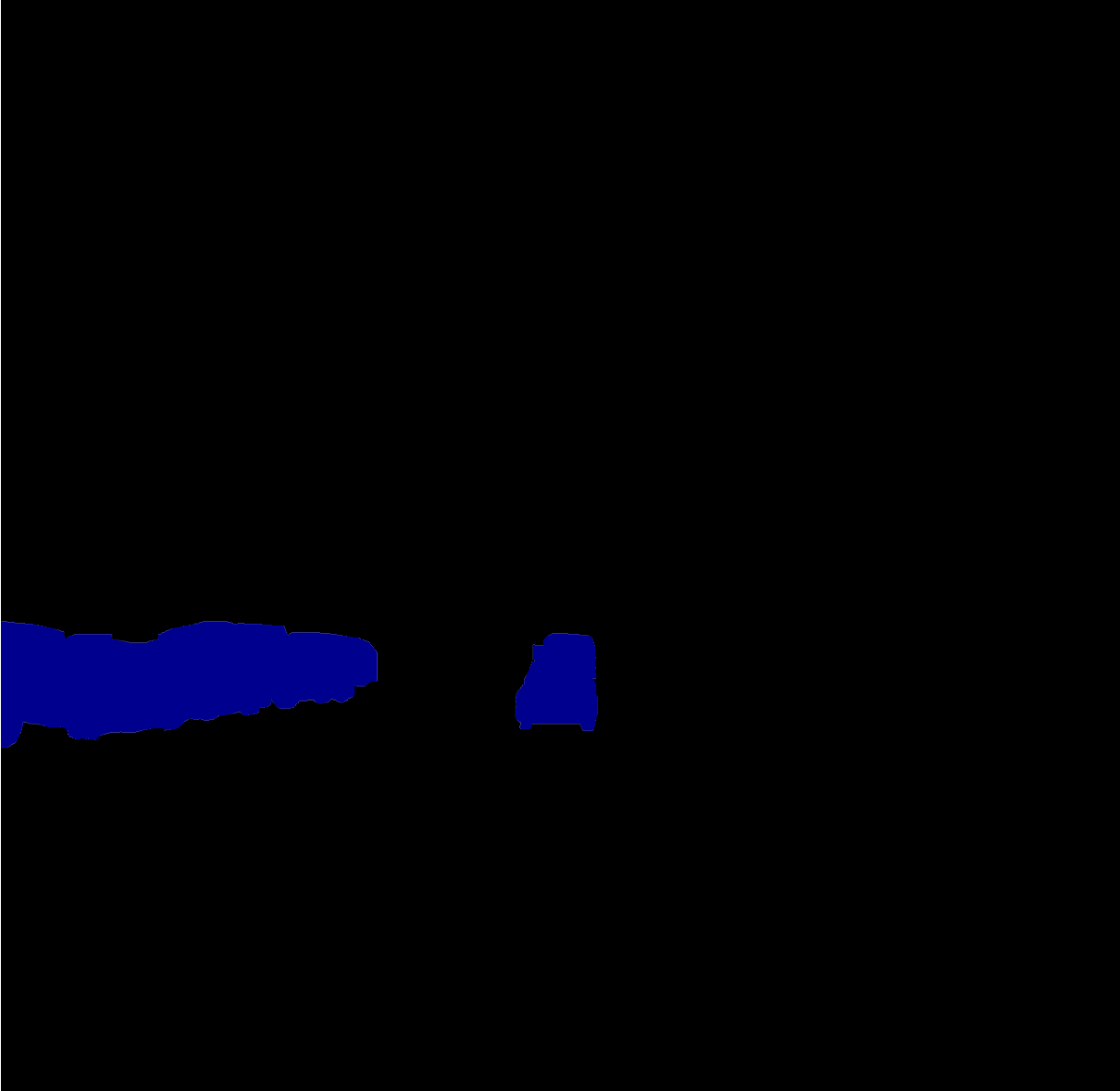}
    \end{minipage}
    \\
    \vspace{5pt}
    \\
    \begin{minipage}{0.02\hsize}
        \raggedright
        {\footnotesize (b)}
    \end{minipage}
    \begin{minipage}{0.1125\hsize}
        \centering
        \includegraphics[width=0.95\hsize, keepaspectratio]{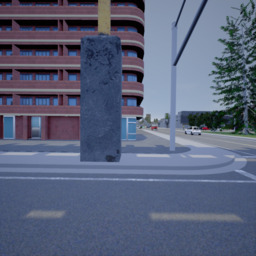}
    \end{minipage}
    \begin{minipage}{0.1125\hsize}
        \centering
        \includegraphics[width=0.95\hsize, keepaspectratio]{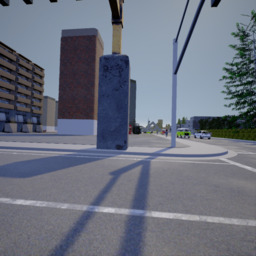}
    \end{minipage}
    \begin{minipage}{0.1125\hsize}
        \centering
        \includegraphics[width=0.95\hsize, keepaspectratio]{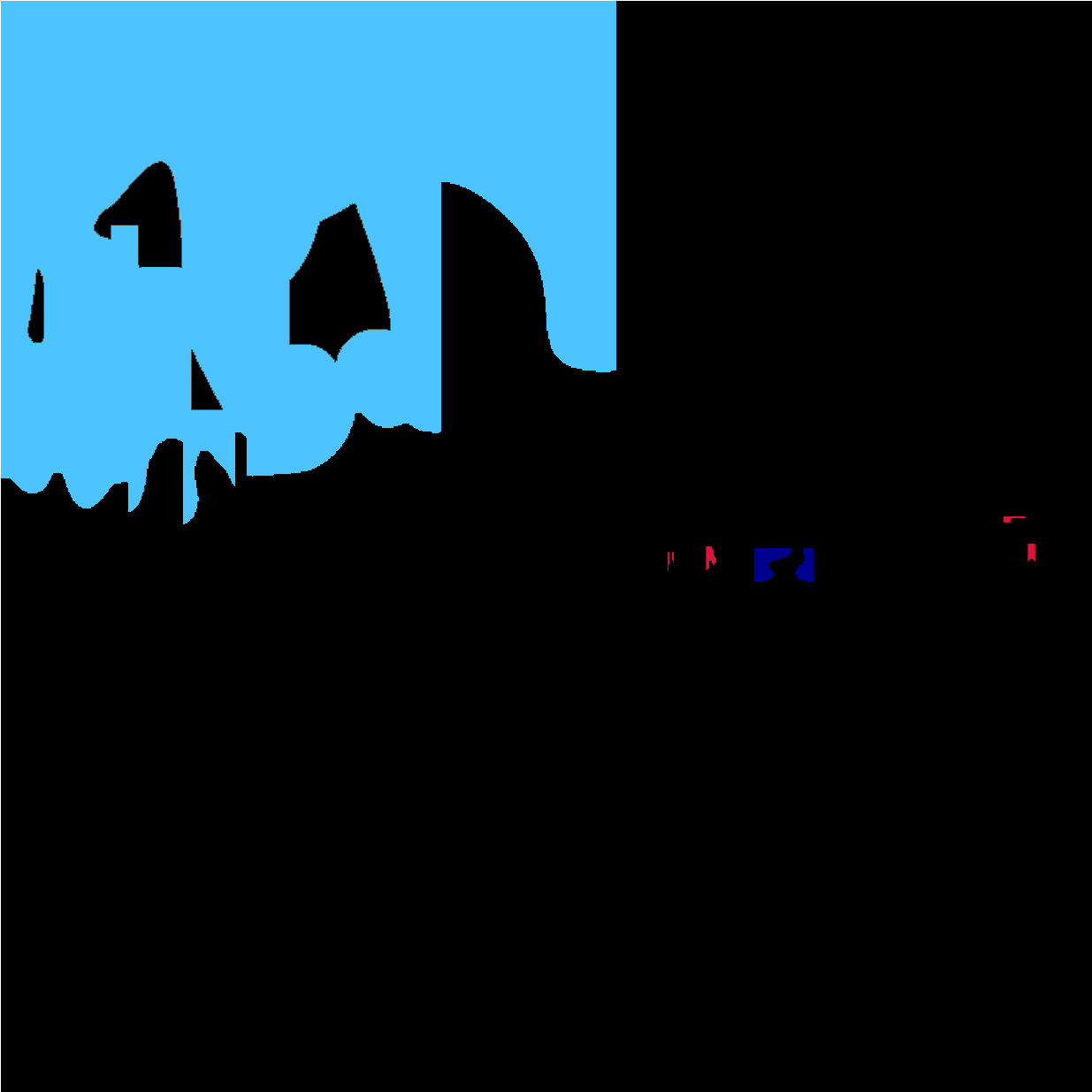}
    \end{minipage}
    \begin{minipage}{0.1125\hsize}
        \centering
        \includegraphics[width=0.95\hsize, keepaspectratio]{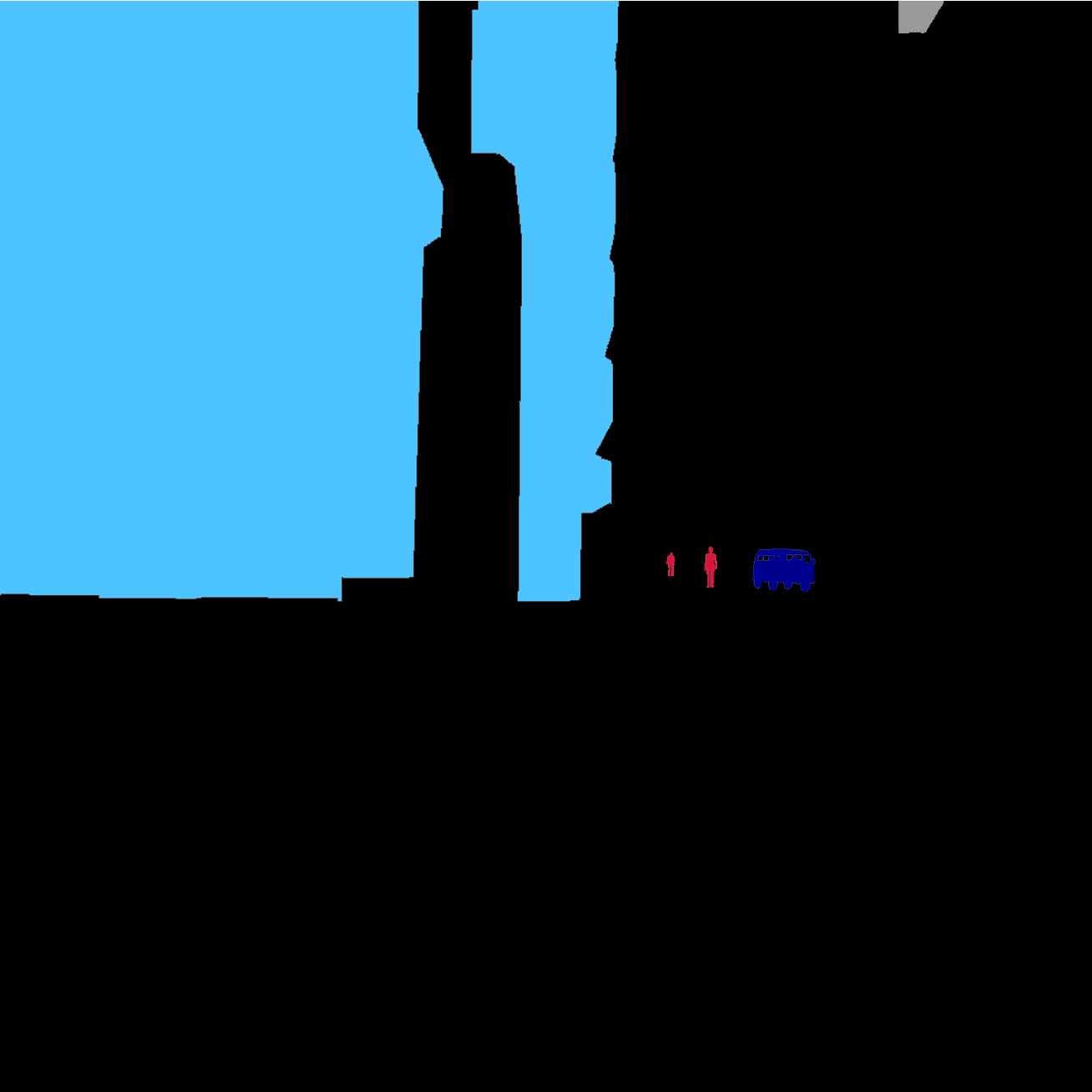}
    \end{minipage}
    \hfill%
    \begin{minipage}{0.02\hsize}
        \raggedright
        {\footnotesize (d)}
    \end{minipage}
    \begin{minipage}{0.1125\hsize}
        \centering
        \includegraphics[width=0.95\hsize, keepaspectratio]{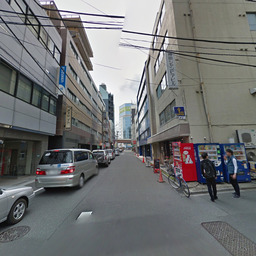}
    \end{minipage}
    \begin{minipage}{0.1125\hsize}
        \centering
        \includegraphics[width=0.95\hsize, keepaspectratio]{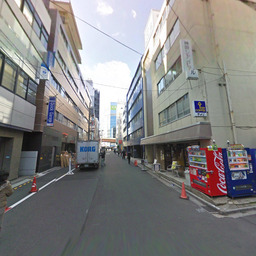}
    \end{minipage}
    \begin{minipage}{0.1125\hsize}
        \centering
        \includegraphics[width=0.95\hsize, keepaspectratio]{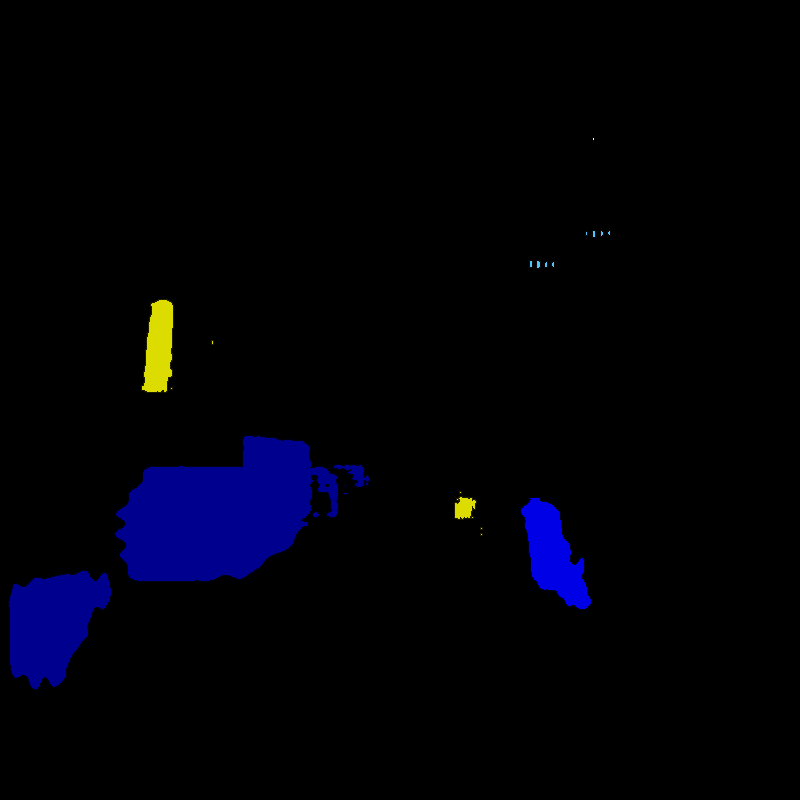}
    \end{minipage}
    \begin{minipage}{0.1125\hsize}
        \centering
        \includegraphics[width=0.95\hsize, keepaspectratio]{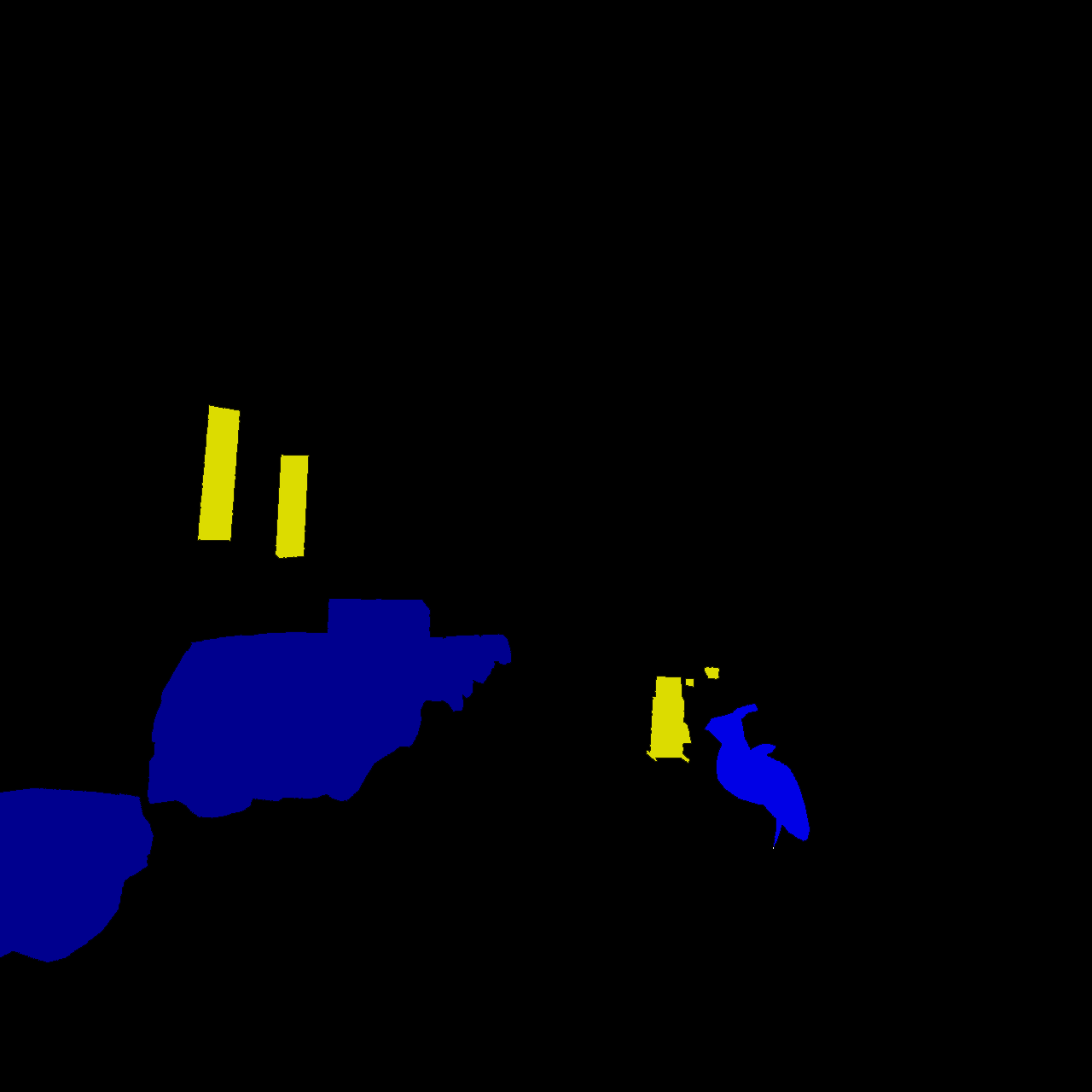}
    \end{minipage}
    \\
    \\
    \begin{minipage}{0.02\hsize}
        \ 
    \end{minipage}
    \begin{minipage}[t]{0.1125\hsize}
        \centering
        {\footnotesize $t_0$}
    \end{minipage}
    \begin{minipage}[t]{0.1125\hsize}
        \centering
        {\footnotesize $t_1$}
    \end{minipage}
    \begin{minipage}[t]{0.1125\hsize}
        \centering
        {\footnotesize Output}
    \end{minipage}
    \begin{minipage}[t]{0.1125\hsize}
        \centering
        {\footnotesize Ground Truth}
    \end{minipage}
    \hfill%
    \begin{minipage}{0.02\hsize}
    \end{minipage}
    \begin{minipage}[t]{0.1125\hsize}
        \centering
        {\footnotesize $t_0$}
    \end{minipage}
    \begin{minipage}[t]{0.1125\hsize}
        \centering
        {\footnotesize $t_1$}
    \end{minipage}
    \begin{minipage}[t]{0.1125\hsize}
        \centering
        {\footnotesize Output}
    \end{minipage}
    \begin{minipage}[t]{0.1125\hsize}
        \centering
        {\footnotesize Ground Truth}
    \end{minipage}
    \caption{Qualitative examples of object-based change detection network (OBJ-CDNet). (a), (b) Examples from CARLA-OBJCD dataset. (c), (d) Examples from GSV-OBJCD dataset.}
    \label{fig:qualitative_examples}
\end{figure*}

% ==========================
% Experiments
% ==========================
\section{Experiments}
\label{sec:experiments}

In this section, we describe the two experimental results, both quantitative and qualitative
on the synthetic and real image datasets.

\subsection{Implementation Details}

During training, we construct object graph from the detected objects that best match ground truth objects based on the IoU between the bounding boxes. For change detection network, we assume correct matching, that is, we calculate the contrastive loss of \cref{eq:contrastive_loss} for the object pairs in ground truth matching. We set the margin parameter $\tau_m$ of the loss as $0.1$.
As the optimization method, Adam \cite{Kingma2015} is used  with an initial learning rate of $0.001$ for the object detection network, $0.0005$ for the graph matching network, and $0.005$ for the change detection network.

During inference, an object pair is classified as `changed' when the feature distance $d$ in \cref{eq:contrastive_loss} is larger than $0.05$.
For the graph matching network, the confidence threshold $\gamma$ in \cref{eq:matching_inf} is set as $0.5$.

We used Faster R-CNN \cite{ren2015faster} with ResNet-101 \cite{he2016deep} as the object detection network, which is pre-trained for $10$ epochs on the Cityscapes dataset \cite{Cordts2016Cityscapes} and then trained for $15$ epochs using each target dataset of GSV-OBJCD for fine-tuning. 
Moreover, we set the hyperparameters as $d = 0.5$, $\mu_{\bf D} = 2$, $z = 0.1$, and $\tau=50$.

The pre-training of the OBJ-CDSegNet requires instance mask annotations in addition to bounding box annotations.
Since our CARLA- and GSV-OBJCD dataset only have bounding box annotations,
we prepare datasets that have both bounding box and instance mask annotations
for pre-training of OBJ-CDSegNet.
For the pre-training of the network on CARLA-OBJCD dataset,
we newly rendered images with bounding box and instance mask annotations using Carla simulator.
For the pre-training of the network on GSV-OBJCD dataset,
we used the Cityscapes dataset \cite{Cordts2016Cityscapes}.
We pre-train OBJ-CDSegNet for 15 epochs on the CARLA-OBJCD-rainy dataset
and 20 epochs on the Cityscapes dataset.
We use Adam as the optimizer and set the learning rate to $1.0 \times 10^{-5}$
for the object detection network to avoid a decrease in the performance of Mask R-CNN.
For other parameters, the same values as OBJ-CDNet are used.

% ==========================
% Evaluation of Graph Matching
% ==========================
\subsection{Evaluation of Graph Matching}
\label{subsec:ablation}

To analyze the effectiveness of deep graph matching network and our proposed epipolar constraint,
we conduct detailed experiments on the object matching network with the synthetic dataset.

\subsubsection{Dataset}

We use CARLA-OBJCD synthetic dataset, which is described in \cref{subsec:carla_dataset},
with ground truth object bounding boxes and correspondences.
We split the image pairs into $10,000/5,000$ for training/testing sets.
To correctly evaluate generalization error, we use different maps of CARLA simulator for rendering the training and testing images.
To analyze the robustness to the viewpoint difference, four different test sets with various viewpoint differences ($\#1 \sim \#4$ described in \cref{subsec:carla_dataset}) are used for evaluation.

\subsubsection{Metrics}

To test the performance of graph matching, the following metric is used as an accuracy of matching:
\begin{equation}
  \text{Accuracy} = \frac{\sum_{p \in P} [ \, p = p^{gt} \,]}{|P|}
\end{equation}
where $p^{gt}$ are the ground-truth matched pairs.
Let $P$ be the set of matched pairs of graph $G_1$ and $G_2$, and $\phi$ the pairs with no matched nodes.
For instance, the pair with no matched nodes $i \in G_1$ can be described as $p = \{i, \, \phi \}, \, p^{gt} = \{i, \, \phi \}$. 
Then if $p = p^{gt}$, $\left[ p = p^{gt} \right] = 1$, otherwise $0$.

\subsubsection{Baselines and proposed method}

We use the following methods as baselines; NN: nearest neighbor matching on deep node features, 
ENN: a variant of NN that is weighted by the epipolar distance obtained by \cref{eq:epi_dist}, and GMN: the graph matching network proposed in \cite{Zanfir_2018_CVPR}.
For the proposed methods, we build two variants of EGMNet: EGMNet (FC) and EGMNet (DT) that respectively use fully-connected and Delaunay object graphs.

\subsubsection{Results}

\Cref{tab:graph_matching_result} shows the object matching accuracies of the baseline methods and ours.
For the CARLA-OBJCD synthetic dataset, there is not much difference in their performance when each input image pair has the same viewpoint (set \#1).
However, the larger the camera pose difference is, the more the matching accuracies of the baseline methods deteriorate.
On the other hand, the proposed EGMNet (FC/DT) can keep its accuracy even for large camera pose differences, which confirm the robustness of the method against viewpoint differences.
Moreover, EGMNet (FC/DT) outperforms other baseline methods for the GSV-OBJCD real dataset.

Comparing the result of ENN with NN, we can tell that the epipolar constraint plays a significant role in achieving robust matching. Moreover, the use of graph structure also improves matching performance, showing that the relationship between objects is an important cue for the matching.

The graph structure of objects in the scene is not typically apparent. 
Hence, we compared Delaunay and fully-connected graphs
where we defined nodes as the center of the rectangular area.
As shown in \cref{tab:graph_matching_result},
EGMNet using the Delaunay graph achieved higher matching accuracy than with a fully-connected one.
This result shows that the matching accuracy depends on the geometric relationship of nodes.

% ==========================
% Evaluation of Pixel-wise Change Detection
% ==========================
\subsection{Evaluation of Pixel-wise Change Detection}

The second experiment is the evaluation of pixel-wise change detection accuracy, where we compare our attention- and segmentation-based method with fully-supervised baseline methods.

\subsubsection{Dataset}

We use CARLA- and GSV-OBJCD dataset,
which is described in \cref{subsec:carla_dataset,,subsec:gsv_dataset} respectively,
with ground truth object bounding boxes, correspondences, and change masks.
The number of samples for training and testing sets are set as $10,000/5,000$ for CARLA-OBJCD dataset and 400/100 for GSV-OBJCD dataset.
As we do not have much samples in the GSV-OBJCD dataset, we perform five-fold cross-validation for the dataset.

\subsubsection{Baseline}

As the baseline methods,
we chose CDNet \cite{sakurada2017dense}, CSCDNet \cite{sakurada_icra2020} and CosimNet-3layer-l2 \cite{guo2018learning},
which are state-of-the-art methods for pixel-wise scene change detection.
We train them using the ground-truth of the object bounding boxes and instance mask for all the experiments.
For the ablation study of object detection, we compare results with and without the ground-truth.

\subsubsection{Metrics}

As in previous work \cite{sakurada2017dense, sakurada_icra2020},
we report the mean intersection-over-union (mIoU) of each method for CARLA- and GSV-OBJCD dataset.

\subsubsection{Results}

\Cref{tab:change_detection_pixel_carla_result,,tab:change_detection_pixel_gsv_result} show the pixel-wise change detection accuracy of each method.
The scores in the brackets are the values
when the ground truth boxes are used instead of the estimated ones to construct the object graphs
in our proposed network.
Where camera viewpoint differences cause significant performance degradation of the pixel-based baseline methods, OBJ-CDNet and OBJ-CDSegNet can sustain their performance.
As~\cite{guo2018learning,sakurada_icra2020,sakurada2017dense} predict pixel-level change maps, the more viewpoint changes increase, the more appearance changes increase.
However, it is challenging to distinguish these two types of changes based on local features so the error of the change detection increases.
Conversely, our method trains graph matching based on bounding boxes, its RoI features, and soft geometric constraints (normalized epipolar distance).
% それに対し、本提案手法では、物体のbounding boxとその特徴量をgraph matchingベースで学習を行い、さらにsoft geometric constrain (normalized epipolar distance)を用いて誤対応を除去しているため、視点の変化に対しよりロバストになっている。
The ground-truth of object bounding boxes gives much higher accuracy than object detection, which shows a potential for performance improvement of our method with a more accurate object detection method.
% Moreover, a limitation of OBJ-CDNet is the coarseness of the change attention map, which causes estimation errors of pixel-wise change.
% It can be improved by utilizing a semantic segmentation netowrk, which can be pre-trained from standard semantic segmentation datasets. 

This result also shows that the precise pixel-wise segmentation by transfer learning, compared to an object attention map, improves change detection accuracy considerably.
Especially when the viewpoint difference is large, OBJ-CDSegNet outperforms two out of three fully-supervised methods in the CARLA-OBJCD dataset experiments.
In GSV-OBJCD dataset experiments, OBJ-CDSegNet attained comparative performance toward fully-supervised methods.
As the result of EGMNet with ground-truth of bounding boxes illustrates, as the accuracy of object detection becomes higher, our network may achieve state-of-the-art performance only by using object-level change annotations.

\Cref{fig:qualitative_examples} shows examples of the semantic change detection for OBJCDNet
with an attention mechanism.
As shown in \cref{fig:qualitative_examples} (a), (b), and (c),
OBJCDNet can accurately detect scene changes even without the change mask annotations.
On the other hand, in \cref{fig:qualitative_examples} (d) the change mask of the small objects is incorrect.
This result shows the limitations of our proposed method.

\section{Conclusion\label{sec:conclusion}}
This paper proposes an object-based change detection network (OBJ-CDNet) based on an epipolar-guided graph matching network (EGMNet). 
To the extent of our knowledge, this is the first work to perform object-based scene change detection using an end-to-end deep learning approach, and the first to introduce the epipolar constraint to a graph matching network.
Furthermore, we created the first publicly available large-scale dataset to benchmark scene change detection.
Our experiments and ablation studies show not only the effectiveness of our approach but also the potential of object-based scene change detection.

\section*{Acknowledgment}
This work was supported by JSPS KAKENHI Grant Number 20H04217.

% clearpage\mbox{}Page \thepage\ of the manuscript.
% clearpage\mbox{}Page \thepage\ of the manuscript.

%\clearpage
% ---- Bibliography ----
%
% BibTeX users should specify bibliography style 'splncs04'.
% References will then be sorted and formatted in the correct style.
%
% \bibliographystyle{splncs04}
\bibliographystyle{IEEEtran}
\bibliography{main.bbl}

\begin{thebibliography}{10}
\providecommand{\url}[1]{#1}
\csname url@rmstyle\endcsname
\providecommand{\newblock}{\relax}
\providecommand{\bibinfo}[2]{#2}
\providecommand\BIBentrySTDinterwordspacing{\spaceskip=0pt\relax}
\providecommand\BIBentryALTinterwordstretchfactor{4}
\providecommand\BIBentryALTinterwordspacing{\spaceskip=\fontdimen2\font plus
\BIBentryALTinterwordstretchfactor\fontdimen3\font minus
  \fontdimen4\font\relax}
\providecommand\BIBforeignlanguage[2]{{%
\expandafter\ifx\csname l@#1\endcsname\relax
\typeout{** WARNING: IEEEtran.bst: No hyphenation pattern has been}%
\typeout{** loaded for the language `#1'. Using the pattern for}%
\typeout{** the default language instead.}%
\else
\language=\csname l@#1\endcsname
\fi
#2}}

\bibitem{wang2014cdnet}
Y.~Wang, P.-M. Jodoin, F.~Porikli, J.~Konrad, Y.~Benezeth, and P.~Ishwar,
  ``{CDnet 2014: An expanded change detection benchmark dataset},'' in
  \emph{CVPR Workshop}, 2014.

\bibitem{jhamtani2018learning}
H.~Jhamtani and T.~Berg-Kirkpatrick, ``{Learning to Describe Differences
  Between Pairs of Similar Images},'' in \emph{EMNLP}, 2018, pp. 4024--4034.

\bibitem{Huertas1998}
A.~Huertas and R.~Nevatia, ``{Detecting Changes in Aerial Views of Man-Made
  Structures},'' in \emph{ICCV}, 1998.

\bibitem{bourdis2011constrained}
N.~Bourdis, D.~Marraud, and H.~Sahbi, ``{Constrained optical flow for aerial
  image change detection},'' in \emph{IGARSS}, 2011, pp. 4176--4179.

\bibitem{crispell2012variable}
D.~Crispell, J.~Mundy, and G.~Taubin, ``{A Variable-Resolution Probabilistic
  Three-Dimensional Model for Change Detection},'' \emph{TGRS}, vol.~50, no.~2,
  pp. 489--500, 2012.

\bibitem{Pollard2007}
T.~Pollard and J.~L. Mundy, ``{Change Detection in a 3-d World},'' in
  \emph{CVPR}, 2007, pp. 1--6.

\bibitem{alcantarilla2018street}
P.~F. Alcantarilla, S.~Stent, G.~Ros, R.~Arroyo, and R.~Gherardi,
  ``{Street-View Change Detection with Deconvolutional Networks},''
  \emph{Autonomous Robots}, vol.~42, no.~7, pp. 1301--1322, 2018.

\bibitem{sakurada2015change}
K.~Sakurada and T.~Okatani, ``{Change Detection from a Street Image Pair using
  CNN Features and Superpixel Segmentation},'' in \emph{BMVC}, 2015.

\bibitem{R-cnn2017MaskR-CNN}
K.~{He}, G.~{Gkioxari}, P.~{Doll^^c3^^a1r}, and R.~{Girshick}, ``{Mask
  R-CNN},'' in \emph{ICCV}, 2017.

\bibitem{Cordts2016Cityscapes}
M.~Cordts, M.~Omran, S.~Ramos, T.~Rehfeld, M.~Enzweiler, R.~Benenson,
  U.~Franke, S.~Roth, and B.~Schiele, ``{The Cityscapes Dataset for Semantic
  Urban Scene Understanding},'' in \emph{CVPR}, 2016.

\bibitem{Dosovitskiy17}
A.~Dosovitskiy, G.~Ros, F.~Codevilla, A.~Lopez, and V.~Koltun, ``{CARLA: An
  Open Urban Driving Simulator},'' in \emph{CoRL}, 2017, pp. 1--16.

\bibitem{panoptic_segmentation}
A.~Kirillov, K.~He, R.~Girshick, C.~Rother, and P.~Dollar, ``{Panoptic
  Segmentation},'' in \emph{CVPR}, 2019.

\bibitem{Radke2005}
R.~J. Radke, S.~Andra, O.~Al-Kofahi, and B.~Roysam, ``{Image Change Detection
  Algorithms: A Systematic Survey},'' \emph{TIP}, vol.~14, no.~3, 2005.

\bibitem{IbrahimEden2008}
D.~C. {Ibrahim Eden}, ``{Using 3D Line Segments for Robust and Efficient Change
  Detection from Multiple Noisy Images},'' in \emph{ECCV}, 2008, pp. 172--185.

\bibitem{daudt2018fully}
R.~C. Daudt, B.~Le~Saux, and A.~Boulch, ``{Fully convolutional siamese networks
  for change detection},'' in \emph{ICIP}, 2018, pp. 4063--4067.

\bibitem{BMVC2015_127}
S.~Stent, R.~Gherardi, B.~Stenger, and R.~Cipolla, ``{Detecting Change for
  Multi-View, Long-Term Surface Inspection},'' in \emph{BMVC}, 2015.

\bibitem{dong20174d}
J.~Dong, J.~G. Burnham, B.~Boots, G.~Rains, and F.~Dellaert, ``{4D Crop
  Monitoring: Spatio-Temporal Reconstruction for Agriculture},'' in
  \emph{ICRA}, 2017.

\bibitem{rosin1998thresholding}
P.~Rosin, ``{Thresholding for change detection},'' in \emph{ICCV}, 1998, pp.
  274--279.

\bibitem{lopez2015features}
F.~J. L{\'o}pez-Rubio and E.~L{\'o}pez-Rubio, ``{Features for stochastic
  approximation based foreground detection},'' \emph{CVIU}, vol. 133, pp.
  30--50, 2015.

\bibitem{wang2018m4cd}
K.~Wang, C.~Gou, and F.-Y. Wang, ``{M4CD: A Robust Change Detection Method for
  Intelligent Visual Surveillance},'' \emph{IEEE Access}, vol.~6, pp.
  15\,505--15\,520, 2018.

\bibitem{Schindler2010}
G.~Schindler and F.~Dellaert, ``{Probabilistic temporal inference on
  reconstructed 3D scenes},'' in \emph{CVPR}, 2010, pp. 1410--1417.

\bibitem{Taneja2011}
A.~Taneja, L.~Ballan, and M.~Pollefeys, ``{Image based detection of geometric
  changes in urban environments},'' in \emph{ICCV}, 2011.

\bibitem{Taneja2013}
A.~Taneja, L.~Ballan, and M.~{Pollefeys}, ``{City-Scale Change Detection in
  Cadastral 3D Models Using Images},'' in \emph{CVPR}, 2013, pp. 113--120.

\bibitem{Sakurada2013}
K.~Sakurada, T.~Okatani, and K.~Deguchi, ``{Detecting Changes in 3D Structure
  of a Scene from Multi-view Images Captured by a Vehicle-Mounted Camera},'' in
  \emph{CVPR}, 2013, pp. 137--144.

\bibitem{MatzenECCV14}
K.~Matzen and N.~Snavely, ``{Scene Chronology},'' in \emph{ECCV}, 2014.

\bibitem{zagoruyko2015learning}
S.~Zagoruyko and N.~Komodakis, ``{Learning to compare image patches via
  convolutional neural networks},'' in \emph{CVPR}, 2015.

\bibitem{khan2017forest}
S.~H. Khan, X.~He, F.~Porikli, and M.~Bennamoun, ``{Forest change detection in
  incomplete satellite images with deep neural networks},'' \emph{TGRS},
  vol.~55, no.~9, pp. 5407--5423, 2017.

\bibitem{guo2018learning}
E.~Guo, X.~Fu, J.~Zhu, M.~Deng, Y.~Liu, Q.~Zhu, and H.~Li, ``{Learning to
  Measure Change: Fully Convolutional Siamese Metric Networks for Scene Change
  Detection},'' \emph{Arxiv}, 2018.

\bibitem{daudt2018high}
R.~C. Daudt, B.~Le~Saux, A.~Boulch, and Y.~Gousseau, ``{Multitask Learning for
  Large-scale Semantic Change Detection},'' \emph{Computer Vision and Image
  Understanding}, vol. 187, 2019.

\bibitem{sakurada_icra2020}
K.~Sakurada, M.~Shibuya, and W.~Wang, ``{Weakly Supervised Silhouette-based
  Semantic Scene Change Detection},'' \emph{ICRA}, 2020.

\bibitem{park2019robust}
D.~H. Park, T.~Darrell, and A.~Rohrbach, ``{Robust Change Captioning},'' in
  \emph{ICCV}, 2019, pp. 4624--4633.

\bibitem{khan2017learning}
S.~H. Khan, X.~He, F.~Porikli, M.~Bennamoun, F.~Sohel, and R.~Togneri,
  ``{Learning Deep Structured Network for Weakly Supervised Change
  Detection},'' in \emph{IJCAI}, 2017, pp. 2008--2015.

\bibitem{Hamaguchi_2019_CVPR}
R.~Hamaguchi, K.~Sakurada, and R.~Nakamura, ``{Rare Event Detection Using
  Disentangled Representation Learning},'' in \emph{CVPR}, June 2019.

\bibitem{furukawa2020iros}
Y.~Furukawa, K.~Suzuki, R.~Hamaguchi, M.~Onishi, and K.~Sakurada,
  ``{Self-supervised Simultaneous Alignment and Change Detection},'' 2020.

\bibitem{sakurada2017dense}
K.~Sakurada, W.~Wang, N.~Kawaguchi, and R.~Nakamura, ``{Dense Optical Flow
  based Change Detection Network Robust to Difference of Camera Viewpoints},''
  \emph{Arxiv}, 2017.

\bibitem{Zhou2012}
F.~Zhou and F.~D. Torre, ``{Factorized Graph Matching},'' \emph{CVPR}, pp.
  1--15, 2012.

\bibitem{Zanfir_2018_CVPR}
A.~Zanfir and C.~Sminchisescu, ``{Deep Learning of Graph Matching},'' in
  \emph{CVPR}, 2018, pp. 2684--2693.

\bibitem{Zhou2016}
B.~Zhou, A.~Khosla, L.~A., A.~Oliva, and A.~Torralba, ``{Learning Deep Features
  for Discriminative Localization.}'' in \emph{CVPR}, 2016, pp. 2921--2929.

\bibitem{Seo2016ProgressivePrediction}
P.~H. Seo, Z.~Lin, S.~Cohen, X.~Shen, and B.~Han, ``{Progressive Attention
  Networks for Visual Attribute Prediction},'' \emph{BMVC}, pp. 1--19, 2016.

\bibitem{Jetley2018LEARNATTENTION}
S.~Jetley, N.~A. Lord, N.~Lee, and P.~H.~S. Torr, ``{Learn to Pay Attention},''
  \emph{ICLR}, 2018.

\bibitem{ren2015faster}
S.~Ren, K.~He, R.~Girshick, and J.~Sun, ``{Faster R-CNN: Towards real-time
  object detection with region proposal networks},'' in \emph{NeurIPS}, 2015,
  pp. 91--99.

\bibitem{he2016deep}
K.~He, X.~Zhang, S.~Ren, and J.~Sun, ``{Deep Residual Learning for Image
  Recognition},'' in \emph{CVPR}, 2016, pp. 770--778.

\bibitem{Yao2018ExploringVR}
T.~Yao, Y.~Pan, Y.~Li, and T.~Mei, ``{Exploring Visual Relationship for Image
  Captioning},'' in \emph{ECCV}, 2018.

\bibitem{contrastive_loss}
R.~Hadsell, S.~Chopra, and Y.~LeCun, ``{Dimensionality Reduction by Learning an
  Invariant Mapping},'' in \emph{CVPR}, 2006, pp. 1735--1742.

\bibitem{Kingma2015}
D.~Kingma and J.~Ba, ``{Adam: A method for stochastic optimization},'' in
  \emph{ICLR}, 2015.

\end{thebibliography}

\end{document}